\documentclass[10pt,twocolumn, letterpaper]{article}

\usepackage{iccv}
\usepackage{times}
\usepackage{epsfig}
\usepackage{amssymb}

\usepackage[american]{babel}
\usepackage{amsmath}
\usepackage{amsthm}
\usepackage{bm}
\usepackage{booktabs}
\usepackage{algorithm}
\usepackage{algorithmic}
\usepackage[switch]{lineno}
\usepackage{multirow}
\usepackage{listings} 
\usepackage{subfigure}
\usepackage{listings}
\usepackage{wrapfig}
\usepackage{fixltx2e}
\usepackage{amsmath}
\usepackage{bbding}
\usepackage{bbold}

\usepackage{color}
\definecolor{deepblue}{rgb}{0,0,0.5}
\definecolor{deepred}{rgb}{0.6,0,0}
\definecolor{deepgreen}{rgb}{0,0.5,0}
\definecolor{codebrown}{rgb}{0.8,0.44,0.2}
\definecolor{codegray}{rgb}{0.5,0.5,0.5}
\definecolor{codepurple}{rgb}{0.58,0,0.82}
\definecolor{backcolour}{rgb}{0.95,0.95,0.92}
\usepackage[pagebackref=true,breaklinks=true,colorlinks,bookmarks=false]{hyperref}

\def\codefont{\fontfamily{lmtt}\selectfont}
\newcommand{\textcode}[1]{{\normalfont\codefont #1}}

\usepackage{silence}
\iccvfinalcopy % *** Uncomment this line for the final submission

\definecolor{codegreen}{rgb}{0,0.6,0}
\definecolor{codegray}{rgb}{0.5,0.5,0.5}
\definecolor{codepurple}{rgb}{0.58,0,0.82}
\definecolor{backcolour}{rgb}{0.95,0.95,0.92}

\lstdefinestyle{mystyle}{
    backgroundcolor=\color{backcolour},   
    commentstyle=\color{codegreen},
    keywordstyle=\color{magenta},
    numberstyle=\tiny\color{codegray},
    stringstyle=\color{codepurple},
    basicstyle=\ttfamily\footnotesize,
    breakatwhitespace=false,         
    breaklines=true,                 
    captionpos=b,                    
    keepspaces=true,                 
    numbers=left,                    
    numbersep=5pt,                  
    showspaces=false,                
    showstringspaces=false,
    showtabs=false,                  
    tabsize=2
}

\def\iccvPaperID{8688} % *** Enter the ICCV Paper ID here

\ificcvfinal\fi

\begin{document}

\setlength{\abovedisplayskip}{2pt}
\setlength{\belowdisplayskip}{2pt}

%%%%%%%%% TITLE
\title{EMQ: Evolving Training-free Proxies for Automated Mixed Precision Quantization}

\author{Peijie Dong$^{1 \dag}$\quad Lujun Li$^{2 \dag}$\quad Zimian Wei$^{1}$ \quad Xin Niu$^{1}\thanks{Corresponding author, $\dag$ equal contribution. }$\quad Zhiliang Tian$^{1}$\quad Hengyue Pan$^{1}$\\
$^1$ National University of Defense Technology, $^2$ HKUST\\
% Institution1 address\\
{\tt\small $^{1}$\{dongpeijie,  weizimian16,  niuxin,  tianzhiliang,  hengyuepan\}@nudt.edu.cn, $^{2}$lilujunai@gmail.com}
}

\maketitle
% Remove page # from the first page of camera-ready.
% \ificcvfinal\thispagestyle{empty}\fi

%%%%%%%%% ABSTRACT
\begin{abstract}
% after grammarly
Mixed-Precision Quantization~(MQ) can achieve a competitive accuracy-complexity trade-off for models. Conventional training-based search methods require time-consuming candidate training to search optimized per-layer bit-width configurations in MQ. Recently, some training-free approaches have presented various MQ proxies and significantly improve search efficiency. However, the correlation between these proxies and quantization accuracy is poorly understood. To address the gap, we first build the MQ-Bench-101, which involves different bit configurations and quantization results. Then, we observe that the existing training-free proxies perform weak correlations on the MQ-Bench-101. To efficiently seek superior proxies, we develop an automatic search of proxies framework for MQ via evolving algorithms. In particular, we devise an elaborate search space involving the existing proxies and perform an evolution search to discover the best correlated MQ proxy. We proposed a diversity-prompting selection strategy and compatibility screening protocol to avoid premature convergence and improve search efficiency. In this way, our Evolving proxies for Mixed-precision Quantization~(EMQ) framework allows the auto-generation of proxies without heavy tuning and expert knowledge. Extensive experiments on ImageNet with various ResNet and MobileNet families demonstrate that our EMQ obtains superior performance than state-of-the-art mixed-precision methods at a significantly reduced cost. The code is available at https://github.com/lilujunai/EMQ-series.

\end{abstract}

%%%%%%%%% BODY TEXT
\section{Introduction}

% dnn
Deep Neural Networks (DNNs) have demonstrated outstanding performance on various vision tasks \cite{ref01_alexnet, lin2014mscoco}. However, their deployment on edge devices is challenging due to high memory consumption and computation cost \cite{han2015deep}. Quantization techniques \cite{krishnamoorthi2018quantizing, choi2018pact, dong2019hawq} have emerged as a promising solution to address this challenge by performing computation and storing tensors at lower bit-widths than floating point precision, and thus speed up inference and reduce the memory footprint.

\begin{figure}[t]
   \begin{center}
        \includegraphics[width=0.9\linewidth]{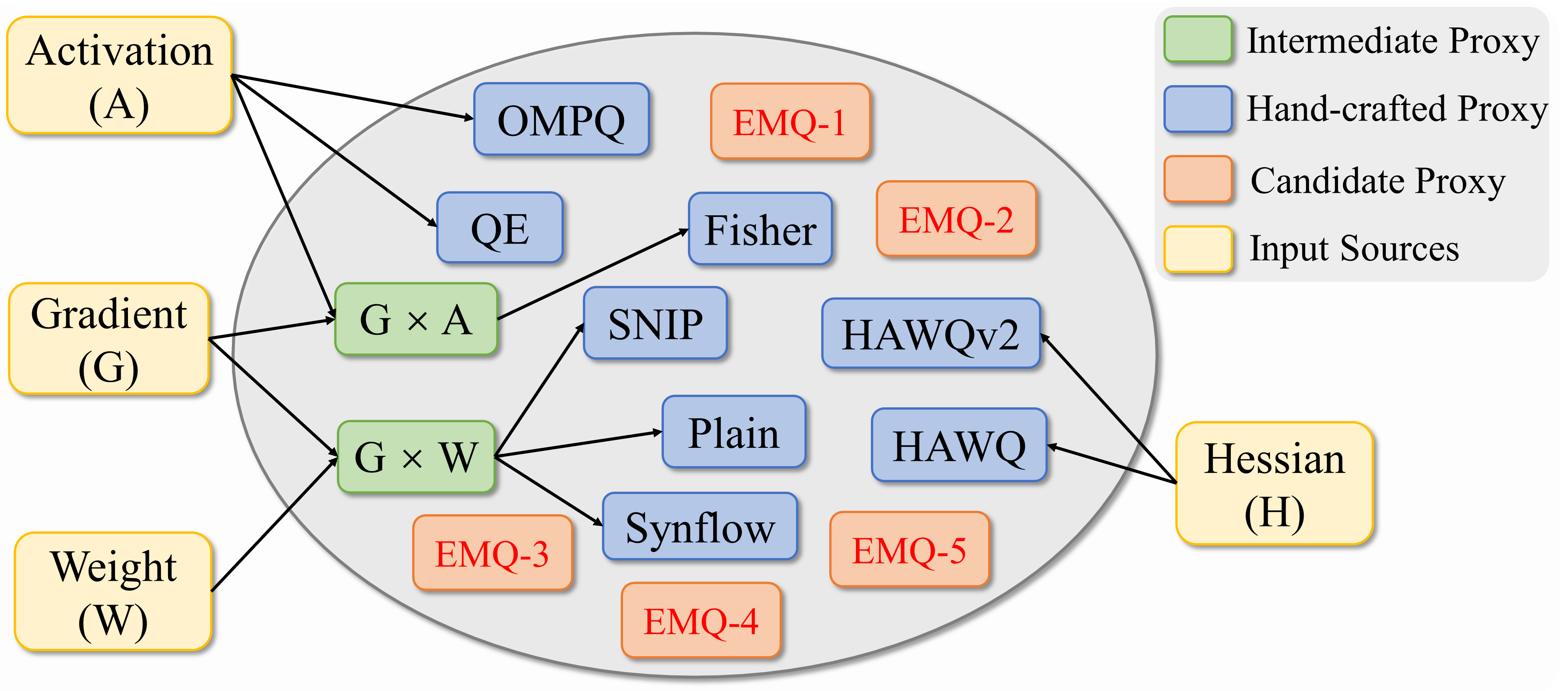}
    \end{center}
    \vspace{-0.5em}
    \caption{Illustration of the search space for EMQ. Our proposed search space encompasses the handcrafted proxies in mixed-precision quantization, whose input sources are activation(A), gradient (G), weight(W), Hessian(H), as well as their combinations (e.g., $G \times W$). The proposed search space highlights the extensive range of possible combinations, emphasizing the significant effort required to discover new MQ proxies.}
    \label{fig:search_space}
    \vspace{-1em}
\end{figure}

% mixed-precision quantization
Mixed-precision quantization (MQ)~\cite{wang2019haq, Jin2019AdaBitsNN,LSQ,guo2020single,dong2019hawq, Habi2020HMQHF} is a technique that assigns different bit-widths to the layers of a neural network to achieve a better accuracy-complexity trade-off and allows for the full exploitation of the redundancy and representative capacity of each layer.
MQ methods can be categorized into training-based and training-free approaches.
\textbf{Training-based methods} for MQ present it as a combinatorial search problem and adopt time-consuming Reinforcement Learning (RL) \cite{wang2019haq}, Evolution Algorithm (EA) \cite{Wang2020APQ}, one-shot \cite{li2021nas}, or gradient-based \cite{wu2018dnas} methods to find the optimal bit-precision setting. 
However, these methods can be computationally intensive and require several GPU days on ImageNet \cite{wang2019haq, Cai2020RethinkingDS}, limiting their applicability in scenarios with limited computing resources or high real-time requirements.
Recently, \textbf{training-free approaches}~\cite{qescore, Ma2021OMPQOM, syflow, dong2019hawq, dong2019hawqv2, Lee2018SNIPSN} have emerged for mixed-precision quantization, which starkly reduces the heavy computation burden. 
These approaches aim to reduce the computational burden by building alternative proxies to rank candidate bit-width configurations. For example, QE~\cite{qescore} uses the entropy value of features to automatically select the bit-precision of each layer. These training-free methods have shown commendable effectiveness in assigning bit-precision to each layer in MQ. 
However, these training-free methods~\cite{dong2019hawq,dong2019hawqv2,yao2021hawq, qescore, Ma2021OMPQOM} have \textbf{two significant limitations}: (i) Lack of correlation analysis between training-free proxies and quantization accuracy. For instance, HAWQ-V2 \cite{dong2019hawqv2}, HAWQ-V3 \cite{yao2021hawq}, and other methods report quantitative results, which couple with quantified strategies and proxies. Thus, it is still unclear whether they can accurately predict the performance of different bit configurations. (ii) The discovery processes for proxies require expert knowledge and extensive trial tuning, which might not fully exploit the potential of training-free proxies. 
These limitations raise \textbf{two fundamental but critical questions}: (1) How can we accurately assess the predictive capability of existing proxies? and (2) How can we efficiently devise new proxies? 

To address the first question, we develop a benchmark, namely, \textbf{MQ-Bench-101}, which comprises numerous bit configurations using the post training quantization strategy. Using this benchmark, we evaluated the performance of several existing training-free proxies, as reported in Tab.~\ref{tab:intro}. Our results demonstrate that the current proxies exhibit limited predictive capabilities. Moreover, we attempt the proxies in training-free NAS and observe that the proxies require bit-weighting for effective quantification \cite{dong2019hawq}. These observations\footnote{There are two routines for proxies in MQ: scoring bit configurations as a whole and evaluating layer-wise sensitivity separately. In this paper, we focus on tackling the former and compare both methods in experiments that are discussed in detail in the App.~\textcolor{red}{D.1}} motivate us to devise improved proxies for MQ.

\begin{figure*}[ht]
    \begin{center}
        \includegraphics[width=0.9\linewidth]{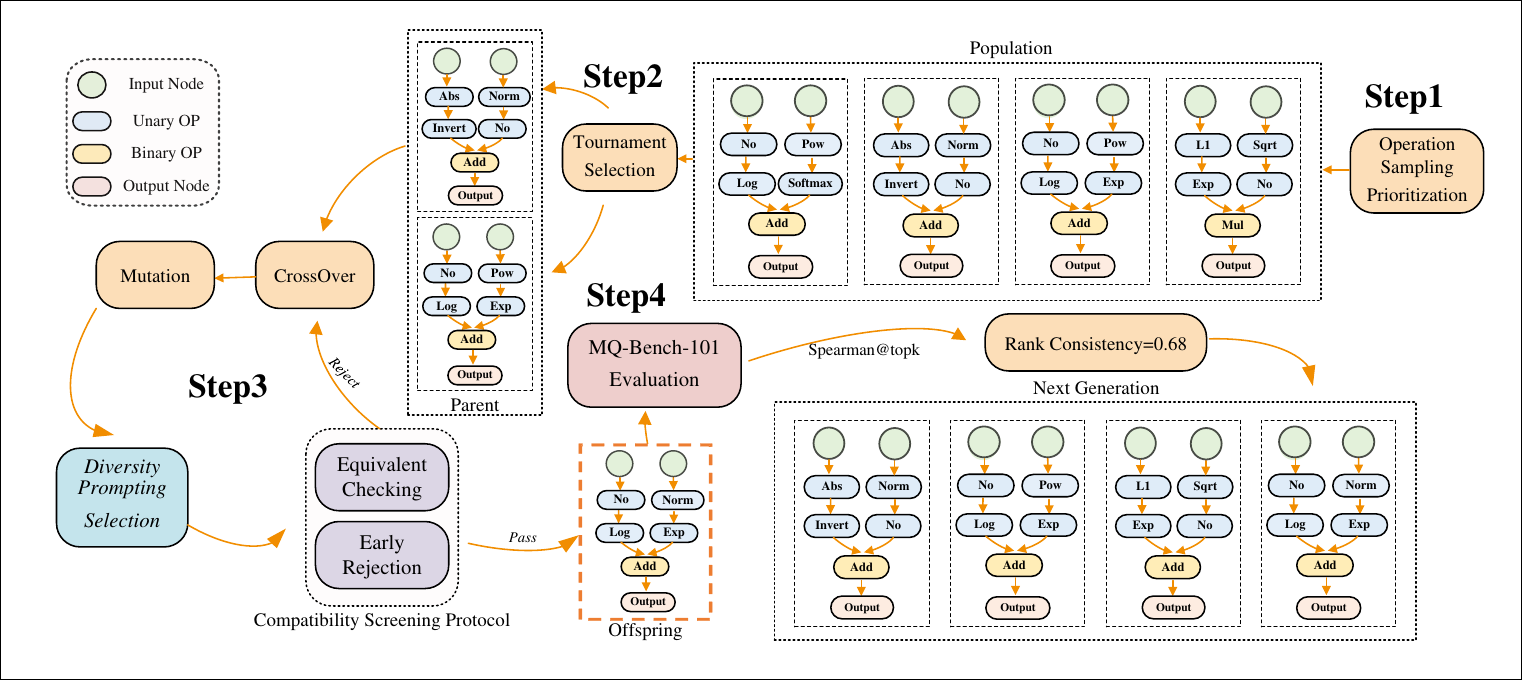}
    \end{center}
    \vspace{-0.5em}
    \caption{Overview of the Evolving training-free proxies for Mixed-precision Quantization (EMQ) framework. The framework involves four main steps: sampling a population of $|\mathcal{P}|$ candidate proxies from the EMQ search space using operation sampling prioritization (Step 1); generating parent proxies through tournament selection (Step 2); producing offspring via crossover, mutation, diversity-prompting selection and compatibility screening protocol (Step 3); and evaluating the offspring on the MQ-Bench-101 to measure the $\mathit{Spearman@topk}$ as the fitness function (Step 4).}
    \label{fig:main_figure}
    \vspace{-1em}
\end{figure*}

As for the second question, we present a general framework, \textbf{Evolving proxies for Mixed-precision Quantization (\textbf{EMQ})}, whose aim is to use a reformative evolving algorithm to automate the discovery of MQ proxies. Specifically, we devise an elaborate and expressive search space encompassing all existing MQ proxies. As shown in Fig.~\ref{fig:main_figure}, we formula MQ proxies as branched computation graphs composed of primitive operations and evolve them according to their predictive ability on MQ-Bench-101.
We notice the importance of the ranking consistency of the top performing bit-widths rather than the overall rank consistency.
To better account for the correlation of the top bit configurations, we introduce $\mathit{Spearman@topk}(\rho_{s@k})$ as the fitness function.
To avoid premature convergence and improve search efficiency of the evolution process, we proposed the diversity-prompting selection strategy and compatibility screening protocol, respectively. We validate our framework on quantization-aware training and post-training quantization tasks. The experiments show that our searched MQ proxy is superior to the existing proxies in predictive capacity and quantization accuracy. 

\vspace{-0.05in} \paragraph{Main Contributions:}
\begin{itemize}\vspace{-0.05in} 
    \item We introduce MQ-Bench-101, the first benchmark for training-free proxies in mixed-precision quantization (Sec.~\ref{sec:mq_bench_101}). \vspace{-0.1in} 
    \item We propose Evolving training-free proxies for Mixed-precision Quantization (EMQ) framework, which includes the diversity-prompting selection to prevent premature convergence and the compatibility screening protocol to improve the evolution search efficiency (Sec.~\ref{sec:emq_framework}). \vspace{-0.1in} 
    \item Experimental results demonstrate the superiority of the searched MQ proxy, indicating the effectiveness and flexibility of our proposed approach (Sec.~\ref{sec:Experiments}). \vspace{-0.1in} 
\end{itemize}

\begin{table}[t]
	\centering
	\small
	\caption{Ranking correlation (\%) of training-free proxies on MQ-Bench-101. The $\mathit{Spearman@topk} (\rho_{s@k})$ are adopted to measure the correlation of the top performing bit configurations on MQ-Bench-101. We reported the mean and std of $\rho_{s@k}$ of 5 runs for all MQ proxies. All implementations are based on the official source code. The 'Time' column indicates the evaluation time (in seconds) for each bit-width configuration.}
	\resizebox{1\linewidth}{!}{
		\begin{tabular}{lrrrr}
			\toprule[1pt]
			Method                              & $\rho_{s@20\%}$    & $\rho_{s@50\%}$    & $\rho_{s@100\%}$   & Time(s) \\
			\midrule
   			BParams                             & 28.67$_{\pm 0.24}$ & 32.41$_{\pm 0.07}$ & 55.08$_{\pm 0.13}$ & 2.59    \\
			HAWQ~\cite{dong2019hawq}            & 23.64$_{\pm 0.13}$ & 36.21$_{\pm 0.09}$ & 60.47$_{\pm 0.07}$ & 53.76   \\
			HAWQ-V2~\cite{dong2019hawqv2}        & 30.19$_{\pm 0.14}$ & 44.12$_{\pm 0.15}$ & 74.75$_{\pm 0.05}$ & 42.17   \\
			OMPQ~\cite{Ma2021OMPQOM}            & 7.88$_{\pm 0.16}$  & 16.38$_{\pm 0.08}$ & 31.07$_{\pm 0.03}$ & 53.76   \\
			QE~\cite{qescore}              & 20.33$_{\pm 0.09}$ & 24.37$_{\pm 0.13}$ & 36.50$_{\pm 0.06}$ & 2.15    \\
			SNIP~\cite{Lee2018SNIPSN}            & 33.63$_{\pm 0.20}$ & 17.23$_{\pm 0.09}$ & 38.48$_{\pm 0.09}$ & 2.50    \\
			Synflow~\cite{Tanaka2020PruningNN} & 39.92$_{\pm 0.09}$ & 44.10$_{\pm 0.11}$ & 31.57$_{\pm 0.02}$ & 2.23    \\
			EMQ(Ours)                           & \textbf{42.59}$_{\pm 0.09}$ & \textbf{57.21}$_{\pm 0.05}$ & \textbf{79.21}$_{\pm 0.05}$ & \textbf{1.02}    \\
			\bottomrule[1pt]
		\end{tabular}}
	\label{tab:intro}%
 \vspace{-1em}
\end{table}

\section{Revisiting Training-free Proxies}
\label{revisiting}

Mixed-precision quantization~\cite{wang2019haq,lou2019autoq,wu2018dnas,yu2020search,dong2019hawq,dong2019hawqv2,cai2020zeroq} aims to optimize the bit-width of each layer in a neural network to strike a balance between accuracy and efficiency. To achieve this, the mixed-precision quantization task can be formulated as a search for the best bit-width using training-free proxies. The search objective function is written as the following bi-level optimization form:
\begin{align}
	\min\limits_{\mathcal{Q}}~~ &\mathcal{L}_{val}(\bm{W}^{*}(\mathcal{Q}), \mathcal{Q})\notag\\
	s.t. ~~&\bm{W}^{*}(\mathcal{Q})=\arg \min ~\mathcal{L}_{train}(\bm{W}, \mathcal{Q})\notag\\
	&\Omega(\mathcal{Q})\leqslant \Omega_0
\end{align}
where $\bm{W}$ refers to the quantized network weights, while $\mathcal{Q}$ denotes the quantization policy that assigns different bit-widths to weights and activations in various layers of the network.
The computational complexity of the compressed network with the quantization policy $\mathcal{Q}$ is represented by $\Omega(\mathcal{Q})$. The task loss on the training and validation data is denoted by $\mathcal{L}{train}$ and $\mathcal{L}{val}$, respectively. The resource constraint of the deployment platform is represented by $\Omega_0$.
In order to obtain the optimal mixed-precision networks, the quantization policy $\mathcal{Q}$ and the network weights $\bm{W}(\mathcal{Q})$ are alternatively optimized until convergence or the maximal iteration number. 
However, training-free approaches~\cite{qescore,Ma2021OMPQOM} take different routine. we formula the problem as: 
\begin{align}
    \mathcal{Q}^*=\max_{\mathcal{Q}} \rho(\mathcal{Q}), \mathcal{Q} \in \mathcal{S}
\end{align}
where $\mathcal{Q}^*$ denotes the best MQ proxy in the search space $S$ and $\rho$ denotes the rank consistency of $\mathcal{Q}$. 
Given a neural network of $L$ layers, the MQ proxy can measure the sensitivity of $i$-th layer by $\mathcal{Q}^*(\theta_{i})$. Then, the objective function is:
\begin{align}
    b^{*}=\max _{\mathbf{b}} \sum_{i=1}^L\left(b_i \times \mathcal{Q}^{*}(\theta_j)\right), 
    s.t. \sum_{i=0}^{L} M^{\left(b_i\right)} \leq \Omega_0. 
\end{align}
where $M^{(b_i)}$ denotes the model size of the $i$-th layer under $b_i$ bit quantization and $b^*$ represents the optimal bit-width configuration under the constraint of $\Omega_0$. 

To dive into the design of training-free proxies, we summarizes the existing MQ proxies in Tab.~\ref{tab:revisit}, which include the training-free proxies in neural architecture search~\cite{Lee2018SNIPSN,syflow} and mixed precision quantization proxies~\cite{qescore, Ma2021OMPQOM, dong2019hawq, dong2019hawqv2}. The proxies are categorized based on four types of network statistics as follows:
% Tab.~\ref{tab:revisit} summarizes the MQ proxies, which are conditioned on different inputs as follows:
(1) \textbf{Hessian as input}: HAWQ~\cite{dong2019hawq} employ the highest Hessian spectrum as the MQ proxy in Eqn.~\ref{eqa:hawqv1}, where $H$ is the Hessian matrix and $\lambda_i(H)$ is the $i$-th eigenvalue of $H$. 
HAWQ-V2~\cite{dong2019hawqv2} adopt the average Hessian trace as proxy in Eqn.~\ref{eqa:hawqv2}, where $\mathit{tr}(H_i)$ denotes the trace of $H_i$. (2) \textbf{Activation as input}: OMPQ~\cite{Ma2021OMPQOM} take the activation $\left\{z\right\}_{i}^{N}$ from the the $i$-th layer as input in Eqn.~\ref{eqa:ompq}, where $||\cdot||_F$ denotes the Frobenius norm.
QE~\cite{qescore} take the variance of the activation $\sigma_{\text{act}}^2$ as input in Eqn.~\ref{eqa:qe}, where $C_l$ represents the product of the kernel size $K_l$ and input channel number $C_{l-1}$ for layer $l$. Fisher~\cite{Theis2018FasterGP} take the activation $z$ as input in Eqn.~\ref{eqa:fisher}. (3) \textbf{Gradient as input}: The formula of SNIP~\cite{Lee2018SNIPSN} is shown in Eqn.~\ref{eqa:snip}, where $\mathcal{L}$ is the loss function of a neural network with parameters $\theta$, and $\odot$ is the Hadamard product. Synflow~\cite{Tanaka2020PruningNN} take the weight $\theta$ and gradient $\frac{\partial \mathcal{R}}{\partial \theta}$ as input but do not require any data. (4) \textbf{Weight as input}: Plain~\cite{Mozer1988SkeletonizationAT}, SNIP~\cite{Lee2018SNIPSN}, and Synflow~\cite{syflow} employ the weights as input, as depicted in Eqn.~\ref{eqa:plain}, Eqn.~\ref{eqa:snip}, and Eqn.~\ref{eqa:synflow}. For more related work, please refer to App.~\textcolor{red}{A}. 

\section{EMQ Framework}
\label{sec:emq_framework}

In this section, we devise a search space and detail the specifics of the evolutionary framework. Then, we introduce the EMQ framework and present evidence for the efficiency of the proposed EMQ.

\subsection{EMQ Search Space Design}

To ensure the effectiveness and flexibility of our search space, we devise a comprehensive set of primitives that goes beyond the simple combinations of existing MQ proxies. Our search space comprises four input types, $56$ primitive operations, and three types of computation graphs. We can construct existing MQ proxies by leveraging these elements, as depicted in Fig.~\ref{fig:search_space}. The abundance of operations in our search space enables us to explore a wide range of possible proxies and discover potential ones that previous handcrafted approaches may have overlooked.

\noindent\textbf{Network Statistics as Input.} As depicted in Fig.~\ref{fig:search_space} and Tab.~\ref{tab:revisit}, the EMQ search space incorporates four distinct input types: activation, gradient, weight, and Hessian of convolutional layers, providing a comprehensive foundation of the sensitivity of each layer. Activation represent the feature map of a convolutional layer, while weight denote the weight of each convolutional layer. Gradient and Hessian matrix are the first and second derivatives of the loss function with respect to the convolution parameters, respectively. By combining these inputs with a diverse set of operations, the search algorithm can explore a vast search space and discover novel MQ proxies.

\begin{table}[t]
    \centering
    \caption{Revisiting mainstream handcrafted training-free proxies for mixed-precision quantization. The proxies are categorized based on four types of network statistics: Hessian matrix (denoted as ``H"), activation (denoted as ``A"), gradient (denoted as ``G"), and weights (denoted as ``W").}
    \resizebox{\linewidth}{!}{
        \begin{tabular}{c|c|c}
            \toprule[1pt]
            Type                                       & MQ Proxy                                 & Formula                                      \\ \midrule[0.9pt]
            \multicolumn{1}{c|}{\multirow{2}{*}{H}}    & HAWQ~\cite{dong2019hawq}              & $\begin{aligned}\label{eqa:hawqv1}\mathit{spectrum}(H) = \max_i \{\lambda_i(H)\} \end{aligned}$  \\ \cline{2-3}
            \multicolumn{1}{c|}{}                      & HAWQ-V2~\cite{dong2019hawqv2}            & $\begin{aligned}\label{eqa:hawqv2}\mathit{trace}(H) = \frac{1}{n}\sum_{i=1}^n \mathit{tr}(H_i)\end{aligned}$  \\ \midrule

            \multicolumn{1}{c|}{\multirow{2}{*}{A}}    & OMPQ~\cite{Ma2021OMPQOM}                & $\begin{aligned}\label{eqa:ompq}\mathit{orm}(z) = \frac{||z_j^T z_i||^2_{z}}{||z_i^T z_i||^2_{z} ||z_j^T z_j||^2_{z}}\end{aligned}$    \\ \cline{2-3}
            \multicolumn{1}{c|}{}                      & QE~\cite{qescore}                 & $\begin{aligned}\label{eqa:qe}\mathit{qe}(\sigma_{\text{act}}) = \sum_{l=1}^L \log\left[\frac{C_l \sigma^2 \sigma_{\text{act}}^2}{\sigma_{\text{act}}^2}\right]+\log(\sigma_{\text{act}}^2)\end{aligned}$      \\ \midrule[0.8pt]
            \multicolumn{1}{c|}{G\&A}                  & Fisher~\cite{Theis2018FasterGP}         & $\begin{aligned}\label{eqa:fisher}\mathit{fisher}(z)=\sum_{z_i \in z}{\left(\frac{\partial \mathcal{L}}{\partial z} z\right)}^2\end{aligned}$  \\ \midrule
            \multicolumn{1}{c|}{\multirow{3}{*}{G\&W}} & Plain~\cite{Mozer1988SkeletonizationAT} & $\begin{aligned}\label{eqa:plain}\mathit{plain}(\theta) = \frac{\partial \mathcal{L}}{\partial \theta} \odot \theta\end{aligned}$   \\ \cline{2-3}
            \multicolumn{1}{c|}{}                      & SNIP~\cite{Lee2018SNIPSN}               & $\begin{aligned}\label{eqa:snip}\mathit{snip}(\theta)=\left|\frac{\partial \mathcal{L}}{\partial \theta} \odot \theta\right|\end{aligned}$    \\ \cline{2-3}
            \multicolumn{1}{c|}{}                      & Synflow~\cite{Tanaka2020PruningNN}      & $\begin{aligned}\label{eqa:synflow}\mathit{synflow}(\theta)=\frac{\partial \mathcal{R}}{\partial \theta} \odot \theta, \mathcal{R}=\mathbb{1}^T\left(\prod_{\theta_i \in \theta}\left|\theta_i\right|\right) \mathbb{1}\end{aligned}$ \\ \bottomrule[1pt]
            % $\begin{aligned}\mathit{orm}(z) = \frac{||z_j^T z_i||^2_{z}}{||z_i^T z_i||^2_{z} ||z_j^T z_j||^2_{z}}\end{aligned}$
        \end{tabular}\label{tab:revisit}
    }
    \vspace{-1em}
\end{table}

\noindent \textbf{Primitive Operations.} 
We employ a set of primitive operations encompassing both unary and binary operations to effectively process the neural network statistics.
To ensure that the search space is sufficiently expressive and encompasses the existing MQ proxies, it is imperative to develop a varied range of operations.
Inspired by AutoML-based methods~\cite{Real2020AutoMLZeroEM, Gu2022AutoLossGMSSG, Liu2021LossFD}, we provide a total of $24$  unary operations and four binary operations to form the EMQ search space. Since the intermediate variables can be scalar or matrix, the total number of operations is $56$.
To efficiently aggregate information from different types of input, we propose aggregation functions to produce the final scalar output of the computation graph.
The opulence of operations in the EMQ framework serves as the cornerstone to construct a diverse and expressive search space that can effectively capture the essence of MQ proxies and yield high-quality solutions. The Appendix~\textcolor{red}{F} describes all the primitive operations in our search space.

\noindent \textbf{Proxy as Computation Graph.} We present each MQ proxy as a computation graph, which can be classified into three structures: sequential structure, branched structure, and Directed Acyclic Graph (DAG) based structure. The sequential structure is a fundamental computation graph, comprising a sequence of nodes with only one input. The branched structure is a hierarchical data structure composed of only one branch with two inputs, as shown in Fig.~\ref{fig:main_figure}. 
It offers a more potent representational capacity than the sequential structure.
The DAG-based structure is most complex and expressive one, which allows for representing intricate dependencies between nodes. Each intermediate node is computed based on all of its predecessors, making it highly expressive yet complex. However, the intensive computation may suffer from sparsity resulting from dimension incompatibility issue or mathematical errors. Due to the trade-off between expressive ability and complexity, we predominantly utilize branched structure in the EMQ framework. For more details, please refer to the App.~\textcolor{red}{C}.

\noindent\textbf{Sparsity of the Search Space.} We measure the sparsity of a search space using the validity rate metric, which represents the ratio between the number of valid proxies and the total number of sampled proxies.
As shown in Tab.~\ref{tab:validy_rate}, the DAG-based structure achieves a validity rate of only $5.4\%$, indicating the sparsity of this search space. The sparsity can be attributed to the dimension incompatibility problem and the mathematical invalidity, which presents a challenge when searching for an effective proxy for EMQ. The dimension incompatibility issue arises from the fact that the input tensors for each proxy may have different dimensions, which not all operations can accommodate. The mathematical invalidity issue arises due to conflicting requirements of various operations, leading to violations of fundamental mathematical principles in the proxy representation. To enhance the validity rate of the search space and to improve the effectiveness of EMQ, it is crucial to address these challenges.

\subsection{Evolutionary Framework}\label{sec:evolutionary-framework}

Inspired by AutoLoss-Zero~\cite{Li2021AutoLossZeroSL} and AutoML-Zero\cite{Real2020AutoMLZeroEM}, we introduce the Evolving proxies for Mixed-precision Quantization (EMQ) search algorithm. As depicted in Fig.~\ref{fig:main_figure} and Alg.~\ref{alg:evolution}, the EMQ pipeline involves several crucial steps. Firstly, we sample $|\mathcal{P}|$ candidate MQ proxies from the search space via operation sampling prioritization strategy. In each evolution, we select two parent proxies using tournament selections with a selection ratio of $r$. The parents then undergo crossover and mutation with probability $p_c$ and $p_m$, respectively, to produce offspring. To prevent premature convergence, we propose a diversity-prompting selection (DPS) method to introduce diversity into the population and avoid population degradation. We also employ compatibility screening protocol to ensure the quality of the offspring before evaluating them on MQ-Bench-101. We adopt $\mathit{Spearman@topk}$ as the fitness function to better correlate with the top performing bit-widths. Finally, we only preserve the top-performing proxies within the population at each iteration. This process is repeated to identify the promising proxy for $\mathcal{N}$ generations. 

\noindent \textbf{Diversity-prompting Selection} 
To introduce diversity into the population and prevent premature convergence, we implemented a diversity-prompting selection method. Instead of directly adding the offspring into the population, we employ additional random proxies and select the proxy with better performance in the population. There are mainly two benefits: (1) It can explore more candidate proxies with a very small population size and prevent premature convergence. (2) By selecting the best-performing individual among the newly generated individuals and the random individual, the evolution algorithm can converge more quickly and efficiently to an optimal solution. 

\noindent \textbf{Spearman@topk as Fitness}
All individuals are evaluated for rank consistency to determine the fitness function in the EMQ evolutionary algorithm. Intuitively, the correlation of the top-performing bit configurations outweigh the overall rank consistency, because we prioritize the ability to find the optimal bit configuration. 
To address this, we devise the $\mathit{Spearman@topk}$ coefficient, which is based on the vanilla Spearman coefficient but focuses only on the top-$k$ performing bit-widths. 
We denote the number of candidate bit-widths as $M$, the ranking of ground-truth (GT) performance and estimated score (ES) of bit-widths $\{b_i\}_{i=1}^{M}$ are $\{p_i\}_{i=1}^M$ and $\{q_i\}_{i=1}^M$, respectively. 
\begin{equation}
\rho_{s@k} = 1 - \frac{6\sum_{i\in D_k}(p_i-q_i)^2}{k(k^2-1)}
\end{equation}
where $\rho_{s@k}$ is the Spearman coefficient computed on the top-$k$ performing bit-widths based on the GT performance, and $D_k$ is the set of indices of the top-$k$ performing bit-widths based on GT performance $D_k=\{{i|p_i < k\times N}\}$.

\begin{algorithm}[t]
\small
\caption{Evolution Search for EMQ}
\label{alg:evolution}
\textbf{Input}: Search space $\mathcal{S}$, population $\mathcal{P}$, sample ratio $r$, sampling pool $\mathcal{Q}$, top-k $k$, selection ratio $r$, max iteration $\mathcal{N}$. 

\textbf{Output}: \leftline{Best MQ proxy with highest $\rho_{s@k}$.}
\begin{algorithmic}[1]
\STATE Initialize sampling pool $\mathcal{Q}$ := $\emptyset$;
\STATE $\mathcal{P}_0$ := Initialize population$(P_i)$ with SOP; 
\FOR{$i = 1:\mathcal{N}$}
    \STATE Clear sampling pool $\mathcal{Q}$ := $\emptyset$;
    \STATE Randomly select $r \times \mathcal{P}$ subnets $\hat{P_i} \in \mathcal{P}$ to get $\mathcal{Q}$;
    \STATE Candidates $\{A_i\}_{k}$ := GetTopk($\mathcal{Q}$, $k$);
    \STATE \label{line:start} Parent $A_i^1, A_i^2$ := RandomSelect$(\{A_i\}_{k})$;
    \STATE Crossover $A_i^c$ := CrossOver($A_i^1, A_i^2$) with probability $p_c$;
    \STATE Mutate $A_i^m$ := MUTATE($A_i^c$) with probability $p_m$;
    \STATE Randomly sample $A_i^n$ from $S$ with OSP;
    \STATE // Diversity-prompting selection.
    \IF{$\rho_{s@k}(A_i^n) \leq \rho_{s@k}(A_i^m)$}
        \STATE Select $A_i^m$ as offspring $A_i^o$;
    \ELSE
        \STATE Select $A_i^n$ as offspring $A_i^o$;
    \ENDIF 
    \STATE // Compatibility screening protocol.
    \IF{CSP($A_i^o$) is true}
        \STATE Append $A_i^o$ to $P$;
    \ELSE 
        \STATE Perform line \ref{line:start};
    \ENDIF
    \STATE Remove the proxy with the lowest $\rho_{s@k}$;
\ENDFOR
\end{algorithmic}
\end{algorithm}

\noindent \textbf{Compatibility Screening Protocol} To address the sparsity issues, we propose Compatibility Screening Protocol (CSP), which includes the equivalent checking and early rejection strategy. Equivalent checking identify distinct structures that are mathematically equivalent, thereby reducing redundant computation. For branched structure, equivalent checking involves the de-isomorphic process, which employ the Weisfeiler-Lehman Test~\cite{leman1968reduction} to filter out the equivalent structures. For more details, please refer to the App.~\textcolor{red}{D.2}. 
\textbf{The early rejection strategy} aims to efficiently filter out invalid MQ proxies. By leveraging the characteristics of MQ proxies, the early rejection strategy employs meticulous techniques to identify and discard invalid proxies before performing a full evaluation on the MQ-Bench-101.
This strategy significantly reduce the time cost of the evolution process or accelerate the convergence of the evolving algorithm. The early rejection strategy comprises three techniques: sensitivity perception, conflict awareness, and naive invalid check.
\textbf{Sensitivity perception} refers to the ability of a proxy to percept whether it is insensitive to the varying of bit-widths, which denotes the incapable of measuring different bit-width and can be rejected at early stage. 
\textbf{Conflict awareness} allows for the identification of conflicting operations during the search process. For instance, the invert operation is in conflict with itself, as is the revert operation. For more detail please refer to App.~\textcolor{red}{D.3}. 
\textbf{Naive Invalid Check} technique is employed to determine if the estimated score of a proxy is one of $\{-1, 1, 0, \mathit{nan}, \mathit{inf}\}$, indicating that it is indistinguishable. Consequently, such proxies can be rejected at an early stage. For more details, please refer to App.~\textcolor{red}{D.4}.

\noindent \textbf{Operation Sampling Prioritization} 
When searching for MQ proxies, random operation sampling results in a large number of invalid candidates. To mitigate this issue, we propose Operation Sampling Prioritization (OSP), which assigns different probabilities to different operations. For unary operations, we assign a higher probability to the $\mathit{no\_op}$ operation to sparsify the search space. For binary operations, we assign a higher probability to the $\mathit{element\_wise\_add}$ operation to ensure that most cases can function well. The proposed OSP can effectively reduce the number of invalid candidates and improve the efficiency of the search process.

\subsection{Effectiveness of EMQ}

\noindent \textbf{Searched Training-Free Proxy} Here is the formula of the searched MQ proxy:
\begin{equation}
    \mathit{emq}(\theta)=\mathit{log}(|\frac{\partial \mathcal{R}}{\partial \theta}|) \sqrt{\frac{\sum_{i=1}^{n}|\theta_{i}|}{\mathit{numel(\theta)+\epsilon}}}
\end{equation}
where $\mathit{numel}(\theta) = \prod_{i=1}^{n} d_i$ and it denotes the total number of elements in the weight $\theta$, and $d_i$ is the size of the $i$-th dimension. The $\mathcal{R}=\mathbb{1}^T\left(\prod_{\theta_i \in \theta}\left|\theta_i\right|\right) \mathbb{1}$ denotes synaptic flow loss proposed in Synflow~\cite{syflow}.
The input type of proposed proxy is similar to existing MQ proxies~\cite{Lee2018SNIPSN, Mozer1988SkeletonizationAT, syflow}. It comprises two components: the logarithm of the absolute value of the derivative of the scalar loss function $\mathcal{R}$, and the square root of the normalized of the absolute values of the weight $\theta$. 
Table~\ref{tab:intro} illustrates the effectiveness of our proposed EMQ, which outperforms the $\rho_{s@100\%}$ of SNIP~\cite{Lee2018SNIPSN} and Synflow~\cite{syflow} by a substantial margin of $40.73\%$ and $47.64\%\uparrow$, respectively. Additionally, EMQ takes less time to evaluate one bit configuration (about $\times 2$ faster).
\begin{figure}[t]
	\setlength{\abovecaptionskip}{0.cm}
	\setlength{\belowcaptionskip}{-0.cm}
	\centering
	\begin{minipage}[t]{0.50\linewidth}
		\centering
		\includegraphics[width=1\linewidth]{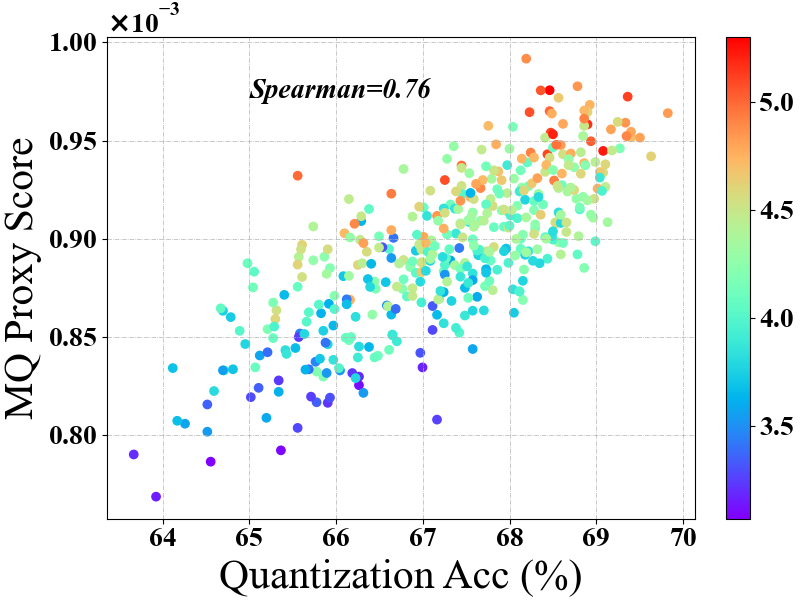}
	\end{minipage}%
	\begin{minipage}[t]{0.50\linewidth}
		\centering
		\includegraphics[width=1\linewidth]{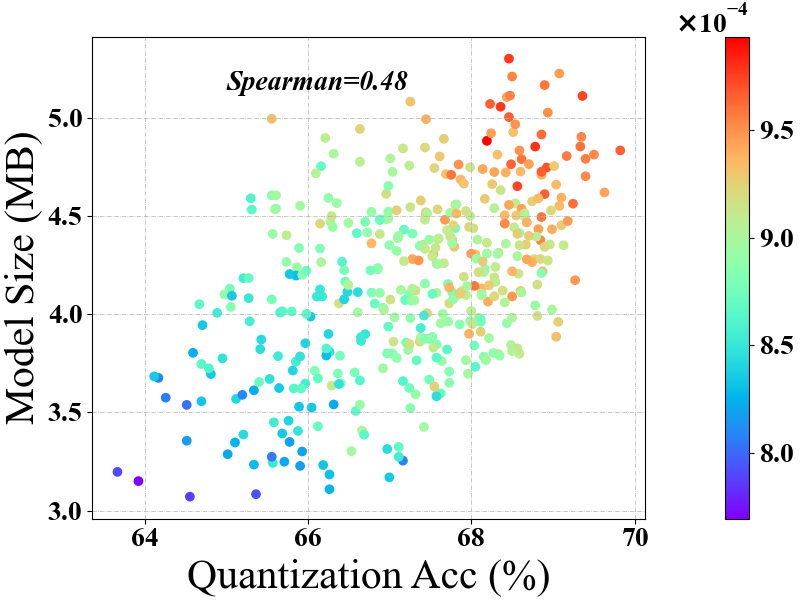}
	\end{minipage}
	\caption{Left: Correlation between the searched EMQ proxy and the quantization accuracy. Right: Correlation between the model size and quantization accuracy.}
	\label{fig:corr}
 \vspace{-1em}
\end{figure}

\noindent \textbf{Correlation of the Searched EMQ Proxy} To evaluate the predictive capability of our searched MQ proxy, we measure the ranking correlation between the searched MQ proxies and the accuracy for bit configurations on MQ-Bench-101. 
The correlation of the searched EMQ proxy with quantization accuracy is exhibited in Fig.~\ref{fig:corr}. 
The figure on the left demonstrates an obvious positive correlation between our searched EMQ method and quantization accuracy, with a Spearman correlation coefficient of $76\%$. The color bar in the figure indicates the corresponding model size of the bit configuration. Conversely, the figure on the right indicates a weak correlation between model size and quantization accuracy, with a Spearman correlation coefficient of only $48\%$. The results suggest that the EMQ proxy has significantly better predictive capability than the baseline (model size as proxy) by a large margin of $28\%\uparrow$.

\begin{figure}[t]
	\setlength{\abovecaptionskip}{0.cm}
	\setlength{\belowcaptionskip}{-0.cm}
	\centering
	\begin{minipage}[t]{0.49\linewidth}
		\centering
		\includegraphics[width=1\linewidth]{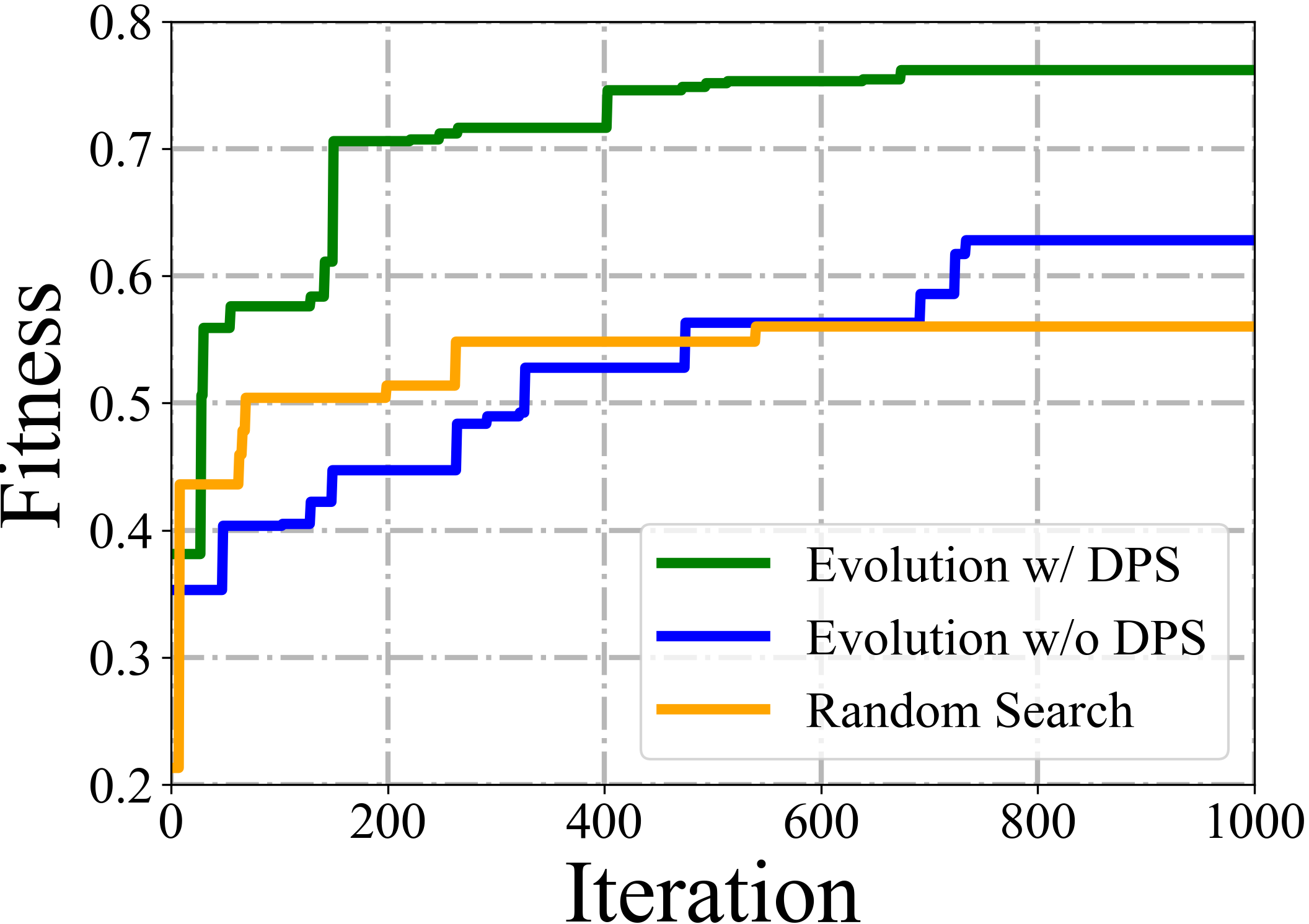}
	\end{minipage}\label{fig:DPS}
	\begin{minipage}[t]{0.49\linewidth}
		\centering
		\centering
		\includegraphics[width=1\linewidth]{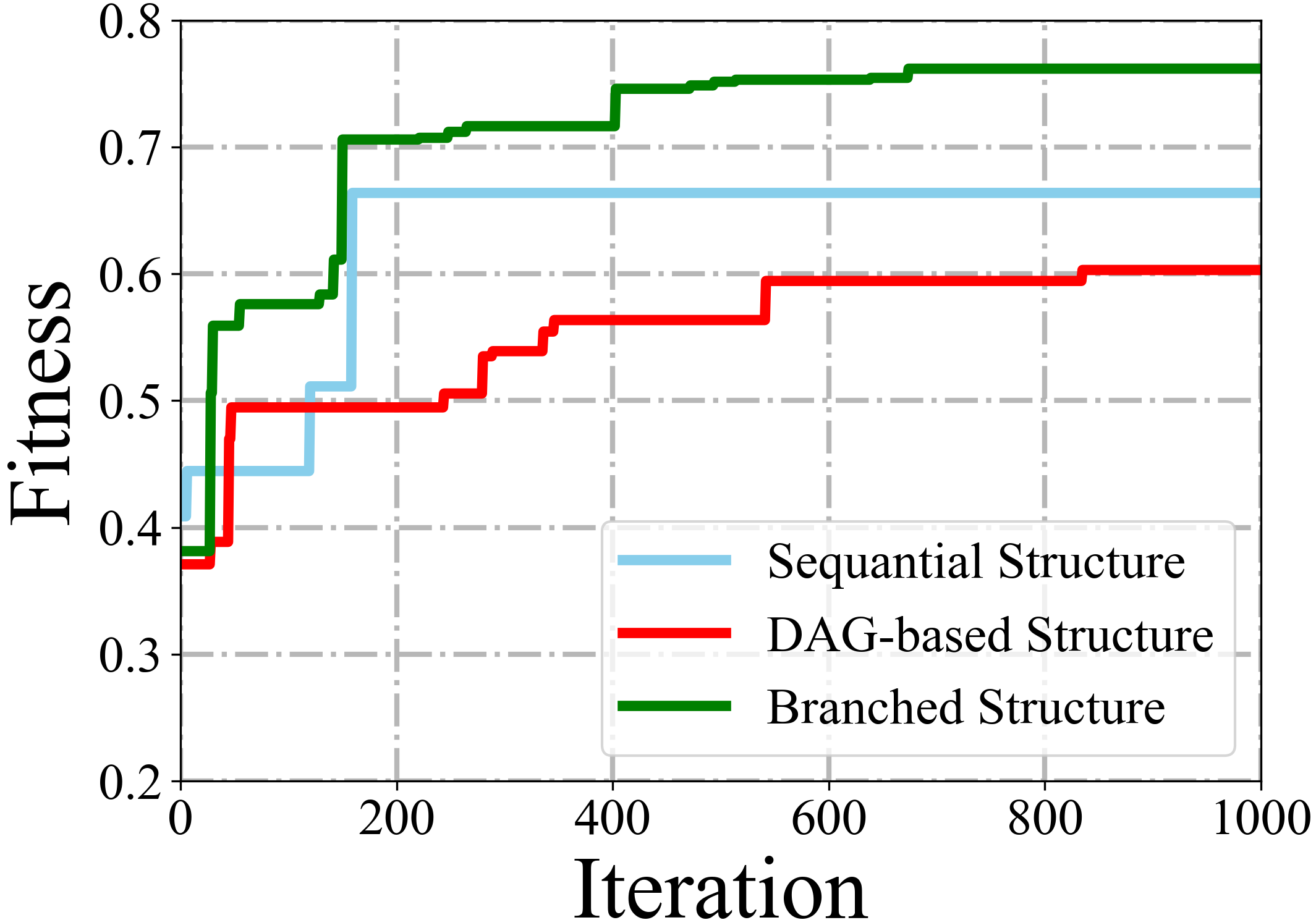}
	\end{minipage}\label{fig:structure}
	\caption{Left: Comparison of the evolutionary search and random search processes, with diversity-prompting selection strategy, denoted as ``DPS". Right: Comparison between sequential, branched, and DAG-based structures during the evolution search.}\label{fig:eff_emq_tree}
 \vspace{-1em}
\end{figure}

\noindent \textbf{Superiority of Branched Structure} 
We present a comparative analysis of the efficiency of three distinct structures: sequential, branched, and DAG-based structure. We assess the validity rate of each search space and investigate the impact of Operation Sampling Prioritization (OSP). 
Tab.~\ref{tab:validy_rate} reveals that the sequential structure has the highest validity rate (41.7\%) due to the simplicity of its computation graph. Nonetheless, this simplicity limits its expressiveness.
The DAG-based structure is theoretically the most expressive search space, but it suffers from a lower validity rate (5.4\%), which leads to slower convergence and higher computational costs. As shown in the right of Fig.~\ref{fig:eff_emq_tree}, we observe that the DAG-based structure fails to achieve better performance, while the sequential structure is trapped in premature convergence due to the lower expressiveness of the search space. In contrast, the branched structure balances expressiveness and computational complexity. With two inputs, the branched structure search space can cover most of the existing MQ proxies and achieve a higher validity rate. For further details, please refer to the App.~\textcolor{red}{C}.

\begin{table}[t]
\centering 
\caption{Validity rate of different search spaces. After applying the operation sampling prioritization strategy, the validity rate of the search spaces is prompted.}
\small
\resizebox{0.8\linewidth}{!}{
\begin{tabular}{lcc}
\toprule[1pt]
\multicolumn{1}{c}{Computation Graph} & w/o OSP (\%) & w/ OSP (\%) \\ \midrule[1pt]
Sequential structure              & 41.70             & 45.85      \\
Branched structure                & 26.40              & 36.45       \\
DAG-based structure               & 5.40              & 6.50         \\ \bottomrule[1pt]
\end{tabular}}\label{tab:validy_rate}
\vspace{-1em}
\end{table}

\section{Experiments} \label{sec:Experiments}

\subsection{Implementation Details}

\noindent \textbf{Datasets} We perform experiments on the ImageNet dataset, which includes $1.2$ million training samples and $50,000$ validation samples. A total of $64$ training samples are randomly selected and the data augmentation techniques used are consistent with those employed in ResNet \cite{resnet}.

\noindent \textbf{Evolution Settings} 
In the evolutionary search process, we employ a population size of $|\mathcal{P}|=20$, and the total number of iteration $\mathcal{N}$ is set to $1000$. The selection ratio $r$ for tournament selection is set to $0.25$, and the probabilities of crossover and mutation, $p_c$ and $p_m$, are set to $0.5$. If the offspring pass the CSP, we randomly sample $50$ bit configurations from MQ-Bench-101 and measure the ranking consistency of the offspring. To determine fitness, we calculated the average of $\rho_{s@20\%},\rho_{s@50\%}$, and $\rho_{s@100\%}$ as the fitness function. During the evolution search, EMQ is extremely efficient, which only needs one NVIDIA RTX 3090 GPU and a single Intel(R) Xeon(R) Gold $5218$ CPU. It only occupies the memory footprint of only one neural network during the evolution process. 

\noindent \textbf{Bit Assignment with Proxy} 
After obtaining the searched EMQ proxy, we employ it to perform bit assignment by selecting the bit configuration with the highest MQ proxy score. Specifically, we first randomly sample a large number of candidate bit-widths that satisfy the model size constraints. We then traverse these candidate bit-widths and select the one with the highest score as the final bit assignment. The process of performing bit assignment is similar to ~\cite{qescore}, and it is extremely fast, taking only a few seconds to evaluate one bit configuration (shown in Tab.~\ref{tab:intro}).

\noindent \textbf{QAT Settings.}
For the QAT experiments, we employed two NVIDIA Tesla V100 GPUs. The quantization framework excludes any integer division or floating point numbers in the network. We set the learning rate to $4e-4$ and the batch size to $512$ for the training process. A cosine learning rate scheduler and SGD optimizer with $1e-4$ weight decay are implemented over 30 epochs. We follow the previous work~\cite{Ma2021OMPQOM} to keep the weight and activation of the first and last layers at $8$ bits, constraining the search space to $\{4,5,6,7,8\}$.

\noindent \textbf{PTQ Settings.}
For the PTQ experiments, we perform them on a single NVIDIA RTX 3090 GPU. We combine EMQ with the BRECQ~\cite{Li2021BRECQPT} finetuning block reconstruction algorithm. In this experiment, we fix the activation precision of all layers to $8$ bits, and limit the search to weight bit allocation in the search space of $\{2,3,4\}$.

\subsection{MQ-Bench-101}
\label{sec:mq_bench_101}

We propose MQ-Bench-101, the first benchmark for evaluating the mixed-precision quantization performance of different bit configurations. To conduct our evaluation, we conduct post training quantization on ResNet-$18$ and assign each layer one of the bit-widths $b=\{2, 3, 4\}$, while keeping the activation to $8$ bits. To manage the computational complexity of the search space, we randomly sample $425$ configurations and attain their quantization performance under post-training quantization settings. MQ-Bench-101 enables us to identify high-performing quantization configurations and compare different MQ proxies fairly. For more details, please refer to the App.~\textcolor{red}{B}.

\subsection{Quantization-Aware Training}
\label{Quantization-Aware Training}

In this experiment, we conducted quantization-aware training on ResNet-$18$/$50$ and compared the results and compression ratios with previous unified quantization methods such as~\cite{park2018value,choi2018pact,zhang2018lq} and mixed-precision quantization methods like~\cite{wang2019haq,chin2020one,yao2021hawq}. The results of our experiments are presented in Tab.~\ref{tab:QAT:resnet18} and Tab.~\ref{tab:qat:res50}.

Our results indicate that EMQ strikes the best balance between accuracy and compression ratio for ResNet-$18$ and ResNet-$50$. For instance, under the bit-width of activation as 6, the searched EMQ proxy achieve a quantization accuracy of $72.28\%$ on ResNet-$18$ with $6.67$Mb and $71$BOPs, which achieves a $0.20\%$ improvement over OMPQ~\cite{Ma2021OMPQOM}. Under the bit-width of activation as 8, EMQ can outperform HAWQ-V3 by $0.75\%$.

Moreover, compared to HAWQ-V3~\cite{yao2021hawq}, EMQ achieve $2.06\%$ higher accuracy while having a slightly smaller BOPs ($71$ vs $72$). 
EMQ achieve an accuracy of $76.70\%$ on ResNet-$50$ with a model size of $18.7$Mb and $148$BOPs, and outperform HAWQ-V3 by $1.31\%$ while having a smaller model size of $17.86$Mb and $148$BOPs compared to $18.7$Mb and $154$BOPs.

\begin{table}
\caption{Mixed-precision quantization results of ResNet-$18$. ``Int" means only including integers during quantization. ``Uni" represents uniform quantization. W/A is the bit-width of weight and activation. $*$ indicates mixed-precision. $\triangledown$ represents not quantizing the first and last layers. ``MS" denotes the model size with bit-parameters and ``BOPs" denotes the bit operations.}
\centering 
\resizebox{82mm}{!}{
\begin{tabular}{cr@{/}lccccc}
    \toprule[1pt]  
    Method                         & W    & A    & Int       & Uni        & MS(M) & BOPs(G) & Top1($\%$)     \\
    \midrule  
    Baseline                       & $32$ & $32$ & \XSolidBrush   & -              & $44.6$          & $1,858$  & $73.09$          \\
    \midrule  
    RVQuant~\cite{Park2018ValueawareQF}                        & $8$  & $8$  & \XSolidBrush   & \XSolidBrush   & $11.1$          & $116$    & $70.01$          \\
    HAWQ-V3~\cite{Yao2020HAWQV3DN}                        & $8$  & $8$  & \CheckmarkBold & \CheckmarkBold & $11.1$          & $116$    & $71.56$          \\
    OMPQ~\cite{Ma2021OMPQOM}                           & $*$  & $8$  & \CheckmarkBold & \CheckmarkBold & $6.7$           & $97$     & $72.30$ \\
    EMQ(Ours)                                          & $*$  & $8$  & \CheckmarkBold & \CheckmarkBold & $6.69$          & $92$  & $\textbf{72.31}$ \\
    \midrule  
    $\text{PACT}^\triangledown$~\cite{choi2018pact}    & $5$  & $5$  & \XSolidBrush   & \CheckmarkBold & $7.2$           & $74$     & $69.80$          \\
    $\text{LQ-Nets}^\triangledown$~\cite{zhang2018lq} & $4$  & $32$ & \XSolidBrush   & \XSolidBrush   & $5.8$           & $225$    & $70.00$          \\
    HAWQ-V3~\cite{Yao2020HAWQV3DN}                        & $*$  & $*$  & \CheckmarkBold & \CheckmarkBold & $6.7$           & $72$     & $70.22$          \\
    OMPQ~\cite{Ma2021OMPQOM}                           & $*$  & $6$  & \CheckmarkBold & \CheckmarkBold & $6.7$           & $75$     & $72.08$ \\
    EMQ(Ours)                                                & $*$   & $6$  & \CheckmarkBold & \CheckmarkBold & $6.69$     & $71$  & $\textbf{72.28}$ \\
    \bottomrule[1pt] 
    \end{tabular}
    }
    \label{tab:QAT:resnet18}
    \vspace{-1em}
\end{table}

\begin{table}
\centering

\caption{Mixed-precision quantization results of ResNet-$50$.}
       \resizebox{82mm}{!}{
\begin{tabular}{cr@{/}lccccc}
    \toprule[1pt]  
    Method                         & W    & A    & Int       & Uni        & MS(M) & BOPs(G) & Top1($\%$)      \\
    \midrule  
    Baseline                       & $32$ & $32$ & \XSolidBrush   & -              & $97.8$          & $3,951$  & $77.72$          \\
    \midrule  
    $\text{PACT}^\triangledown$~\cite{choi2018pact}    & $5$  & $5$  & \XSolidBrush   & \CheckmarkBold & $16.0$          & $133$    & $\textbf{76.70}$ \\
    $\text{LQ-Nets}^\triangledown$~\cite{zhang2018lq} & $4$  & $32$ & \XSolidBrush   & \XSolidBrush   & $13.1$          & $486$    & $76.40$          \\
    RVQuant~\cite{Park2018ValueawareQF}                        & $5$  & $5$  & \XSolidBrush   & \XSolidBrush   & $16.0$          & $101$    & $75.60$          \\
    HAQ~\cite{wang2019haq}                            & *    & $32$ & \XSolidBrush   & \XSolidBrush   & $9.62$          & $520$    & $75.48$          \\
    Onebit-width~\cite{Chin2020OneWB}                    & *    & $8$  & \XSolidBrush   & \CheckmarkBold & $12.3$          & $494$    & $\textbf{76.70}$ \\
    HAWQ-V3~\cite{Yao2020HAWQV3DN}                        & *    & *    & \CheckmarkBold & \CheckmarkBold & $18.7$          & $154$    & $75.39$          \\
    OMPQ~\cite{Ma2021OMPQOM}                           & *    & $5$  & \CheckmarkBold & \CheckmarkBold & $18.7$          & $156$    & $76.28$          \\
    EMQ(Ours)                                          & *    & $5$  & \CheckmarkBold & \CheckmarkBold & $17.86$         & $148$   & $\textbf{76.70}$ \\ 
    \bottomrule[1pt] 
\end{tabular}\label{tab:qat:res50}
\vspace{-1em}
}

\end{table}

\subsection{Post-Training Quantization}
\label{Post-Training Quantization}

\begin{figure}
    \centering
    \includegraphics[width=0.99\linewidth]{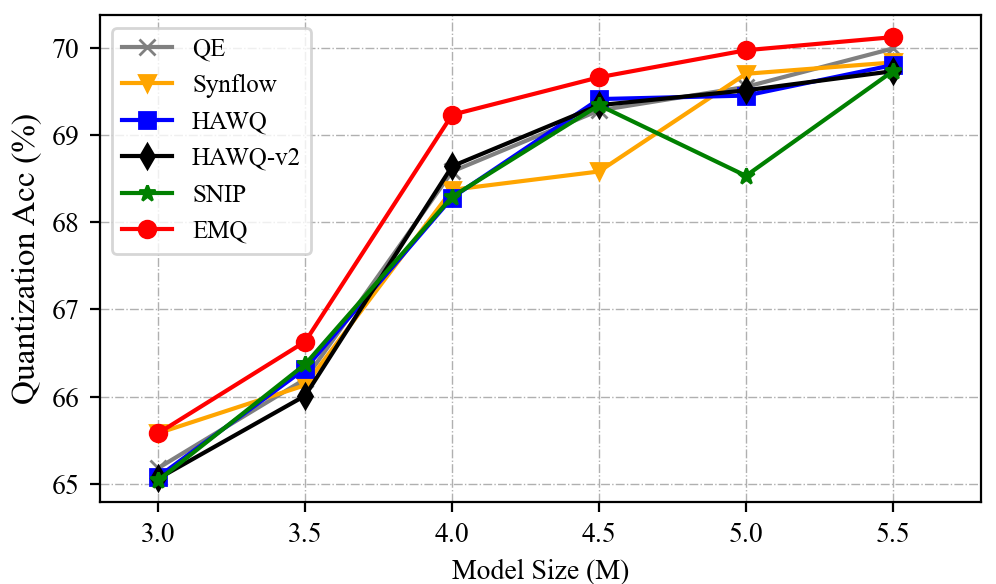}
    \caption{The accuracy and complexity trade-off between MQ proxies and our proposed EMQ approach for ResNet-18.}
    \label{fig:pareto}
\end{figure}
\begin{table}
\centering

 \caption{Mixed-precision post-training quantization results on ResNet-$18$. $\dag$ means using distillation in the finetuning process.}
        \resizebox{82mm}{!}{
	\begin{tabular}{ccccc}
		\toprule[1pt]  
		Method          &
		W/A             &
		Model size(M) &
		Top-1 (\%)      & 
            \#Data         \\
		\midrule  
		Baseline                         & $32$/$32$ & $44.6$ & $71.08$          & -                     \\
		\midrule  
		FracBits-PACT~\cite{choi2018pact}                    & $*$/$*$   & $4.5$  & $69.10$          & $1.2$M             \\
		OMPQ~\cite{Ma2021OMPQOM}                             & $*$/$4$   & $4.5$  & $68.69$          & $64$  \\
        EMQ(Ours)                                & */4    & 4.5  & \textbf{69.66}    & 64   \\ 
		ZeroQ~\cite{cai2020zeroq}                            & $4$/$4$   & $5.81$ & $21.20$          & -                        \\
		$\text{BRECQ}^\dag$ ~\cite{Li2021BRECQPT}             & $4$/$4$   & $5.81$ & $69.32$          & -                       \\
		PACT~\cite{choi2018pact}                             & $4$/$4$   & $5.81$ & $69.20$          & -                     \\
		HAWQ-V3~\cite{Yao2020HAWQV3DN}                          & $4$/$4$   & $5.81$ & $68.45$          & -                     \\
		FracBits-PACT~\cite{choi2018pact}                    & $*$/$*$   & $5.81$ & $69.70$ & $1.2$M            \\
		OMPQ~\cite{Ma2021OMPQOM}                             & $*$/$4$   & $5.5$  & $69.38$          & $64$  \\
        EMQ(Ours)                                & */4   & 5.5   & \textbf{70.12}       & 64    \\
		\midrule  
		BRECQ~\cite{Li2021BRECQPT}                            & $*$/$8$   & $4.0$  & $68.82$          & $1,024$             \\
		OMPQ~\cite{Ma2021OMPQOM}                             & $*$/$8$   & $4.0$  & $69.41$ & $64$ \\ 
        EMQ(Ours)                                & */8   & 4.0   & \textbf{69.92}     & 64 \\ 
		\bottomrule[1pt] 
	\end{tabular}\label{tab:ptq:r18}
 \vspace{-1em}
 }
\end{table}

\begin{table}
\caption{Mixed-precision post-training quantization results on MobileNetV$2$.}
\centering
\resizebox{82mm}{!}{
\begin{tabular}{cccccc}
\toprule[1pt]  
Method & 
W/A & 
Model Size (Mb)&
Top-1 (\%)&
\#Data  \\
\midrule  
Baseline & $32$/$32$ & $13.4$ & $72.49$ & -  \\
\midrule  
BRECQ~\cite{Li2021BRECQPT} & $*$/$8$ & $1.3$ & $68.99$ & $1,024$  \\
OMPQ~\cite{Ma2021OMPQOM} & $*$/$8$ & $1.3$ & $69.62$ & $32$  \\
EMQ(Ours)    &  $*/8$ & $1.3$ & $\textbf{70.72}$  & $64$  \\ 
\midrule  
FracBits~\cite{Yang2020FracBitsMP} & $*$/$*$ & $1.84$  & $69.90$ & $1.2$M  \\
BRECQ~\cite{Li2021BRECQPT} & $*$/$8$ & $1.5$ & $70.28$ & $1,024$  \\
EMQ(Ours)   & $*/8$  & $1.5$  & $\textbf{70.75}$   & $64$  \\ 

\bottomrule[1pt] 
\end{tabular}\label{tab:ptq:r50}
}
\vspace{-1em}
\end{table}
In this experiment, we conduct experiments on ResNet18 and MobileNetV2. 
Our proposed EMQ approach achieves a better trade-off among different model sizes, as illustrated in Tab.~\ref{tab:ptq:r18} and~\ref{tab:ptq:r50}. To achieve this, we adopted the same block reconstruction quantization strategy as OMPQ~\cite{Ma2021OMPQOM}. Our experiments show that under the constraint of model size $\{4.0, 4.5, 5.5\}$, we achieve competitive results, surpassing OMPQ by $0.97\%$, $0.74\%$, and $0.51\%$, respectively.
Moreover, we conducted a series of experiments to evaluate the performance of different model sizes \{$3.0$, $3.5$, $4.0$, $4.5$, $5.0$, $5.5$\} using various quantization proxies, including QE~\cite{qescore}, Synflow~\cite{syflow}, HAWQ~\cite{dong2019hawq}, HAWQ-V2~\cite{dong2019hawqv2}, and EMQ. To strike a trade-off between model complexity and quantization accuracy, we plot the quantization accuracy of each proxy against its respective model size, resulting in a pareto front (as shown in Fig.~\ref{fig:pareto}). The results demonstrate that our EMQ proxy provides a superior trade-off between model complexity and quantization performance when compared to the existing proxies.

\subsection{Ablation Study}
\label{Ablation Study}

As presented in Tab.~\ref{tab:validy_rate}, we observe that the proposed operation sampling prioritization (OSP) technique improves the validity rate of the branched structure by $10.05\%\uparrow$. 
As illustrated in the left of Fig.~\ref{fig:DPS}, diversity-prompting selection (DPS) strategy can indeed prevent premature convergence (Blue line) and outperform the random search baseline (Yellow line) by a large margin. These findings suggest that the OSP and DPS strategy are indispensable components of EMQ.
Tab.~\ref{tab:ablation} demonstrates the effectiveness of the ablation study on improving efficiency through equivalent checking and early rejection when searching for branched structures. By implementing these strategies, we are able to proactively filter out approximately 97\% of failed proxies, resulting in a significant reduction in computational cost.

\begin{table}
	\caption{Efficiency improvement with equivalent checking and early rejection strategy on branched structure.}
	\centering
	\resizebox{\linewidth}{!}{
		\begin{tabular}{c|c|c}
			\toprule[1.pt] 
                Equivalent Checking & Early Rejection & \#Evaluated Proxies \\
			\midrule[1pt]
                \XSolidBrush             & \XSolidBrush                 & $\sim 1 \times 10^4$     \\
			\CheckmarkBold           & \XSolidBrush              & $\sim9 \times 10^3$     \\
			\CheckmarkBold           & \CheckmarkBold    & $\sim 3 \times 10^2$                  \\
			\bottomrule[1pt]
		\end{tabular}\label{tab:ablation}
	}
\end{table}

\subsection{Visualization and Analysis}
\begin{figure}[t]
    \centering
    \includegraphics[width=0.99\linewidth]{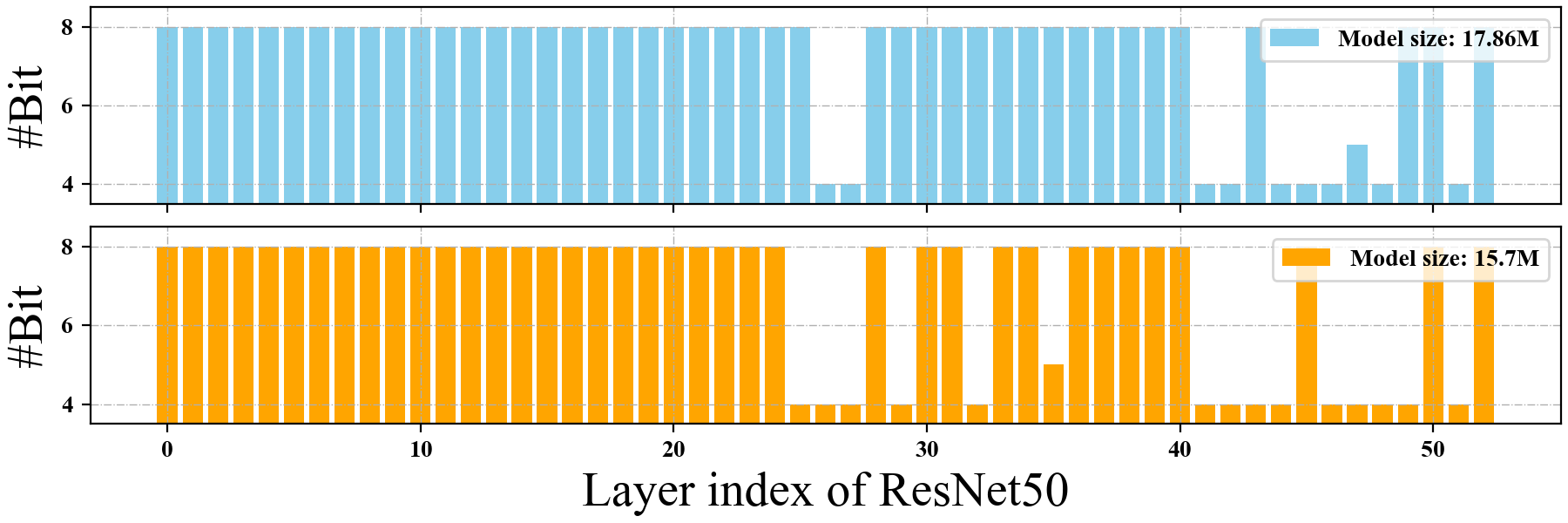}
    \caption{Assignment of bit configurations for weights under $18$M and $16$M model size constraints for ResNet-$50$. The bit-widths are searched for configurations of $\{4, 5, 6, 7, 8\}$}
    \label{fig:best_assign}
    \vspace{-1em}
\end{figure}

To intuitively demonstrate the bit-width assignment generated by the searched EMQ proxy, we visualize the quantization strategy of weights in different layers of ResNet50 with model size constraints of $16$M and $18$M in Fig.~\ref{fig:best_assign}. We observe that for the bit-width assignment under different model constraints, the $29^{th}$, $32^{nd}$, $35^{th}$, and $49^{th}$ layers are assigned lower bit-width, indicating that these layers are not as sensitive as others. Additionally, we can see from the bit-width assignment that the first and last layers have higher bit-width to achieve quantization accuracy.

\section{Conclusion}

In this paper, we present the Evolving proxies for Mixed precision Quantization (EMQ), a novel approach for exploring proxies for mixed-precision quantization (MQ) without requiring heavy tuning or expert knowledge. To fairly evaluate the MQ proxies, we build the MQ-Bench-101 benchmark. We leverage evolution algorithm to efficiently search for superior proxies that strongly correlate with quantization accuracy, using our diversity-prompting selection and compatibility screening protocol. The extensive experiments on the ImageNet dataset on ResNet and MobileNet families demonstrate that our EMQ framework outperforms existing state-of-the-art mixed-precision methods in terms of both accuracy and efficiency. We believe that our work inspires further research in developing efficient and accurate MQ techniques and enables deploying more efficient deep learning models in resource-constrained environments. 

\section{Acknowledgements}
This work is supported by the National Natural Science Foundation of China (No.62025208) and the Open Project of Xiangjiang Laboratory (No.22XJ01012).

\appendix
\part*{Appendix}
In this appendix, we provide additional details and information about EMQ, including related works, the MQ-Bench-101 dataset used for evaluation, computation graph details, and the evolving algorithm. We also provide more information on the routine, Weisfeiler-Lehman test, conflict awareness, naive invalid check, and operation sampling prioritization. Additionally, we analyze the iterations and population of the algorithm and present more ablation studies, including sensitivity analyses of batch size and seeds, as well as primitive operations. The appendix serves as a comprehensive reference for those interested in understanding and implementing the EMQ algorithm.

\section{Additional Related Works}

\subsection{Model Compression}
Model compression techniques have emerged as a crucial research area in deep learning, with the goal of reducing the computational and memory requirements of deep neural networks. The ultimate objective is to make these models more efficient and deployable on resource-constrained devices, such as mobile phones or embedded systems. In the context of computer vision, various compression methods have been proposed, including knowledge distillation~\cite{gou2020knowledge, ickd, liu2023norm, li2023auto, li2022self, shao2023catch, li2022tf}, quantization~\cite{morgan1991experimental,jacob2018quantization}, pruning~\cite{syflow, Tanaka2020PruningNN, Theis2018FasterGP}, and neural architecture search~\cite{liu2018darts, dong2023rd, chen2019pdarts, guo2020single}. Knowledge distillation involves transferring knowledge from a large, accurate teacher network to a smaller, more efficient student network. Quantization reduces the precision of network weights and activations, while pruning removes unnecessary network parameters. Neural architecture search automates the process of designing optimal network architectures for specific tasks. These techniques have demonstrated promising results in reducing the size and complexity of deep neural networks while maintaining or even improving their performance. Unlike other compression techniques, such as pruning or knowledge distillation, which reduce the size of the model by removing parameters, quantization preserves the model architecture and reduces the memory and computation requirements by reducing the precision of the parameters, resulting in faster and more energy-efficient inference.

\subsection{Mixed-Precision Quantization.} 
Quantization~\cite{morgan1991experimental,jacob2018quantization,park2018value,chin2020one} has been widely investigated as an effective technique to accelerate the inference phase of neural networks by converting 32-bit floating-point weight/activation parameters into low-precision fixed-point values. However, the contribution of each layer to the overall performance is to varying extents, and mixed-precision quantization~\cite{wang2019haq,lou2019autoq,wu2018dnas,yu2020search,dong2019hawq,dong2019hawqv2,cai2020zeroq} has been proposed to achieve a better trade-off between accuracy and complexity by assigning different bit-precision to different layers.
Existing mixed-precision quantization methods can be classified into four categories: reinforcement learning-based approaches~\cite{lou2019autoq,wang2019haq,Elthakeb2020ReLeQA}, evolutionary algorithm-based approaches~\cite{Wang2020APQ}, one-shot approaches~\cite{wu2018dnas, Jin2019AdaBitsNN, Habi2020HMQHF} (including differentiable search approaches), and zero-shot approaches~\cite{dong2019hawq, dong2019hawqv2, qescore, Ma2021OMPQOM} (also known as heuristic-based methods). Reinforcement learning-based approaches, such as HAQ \cite{wang2019haq}, use hardware feedback to search the bit-precision in discrete space. Evolutionary algorithm-based approaches, such as APQ~\cite{Wang2020APQ}, jointly search the pruning ratio, the bitwidth, and the architecture of the lightweight model from a hypernet.

\begin{figure}[t]
    \begin{center}
        \includegraphics[width=0.99\linewidth]{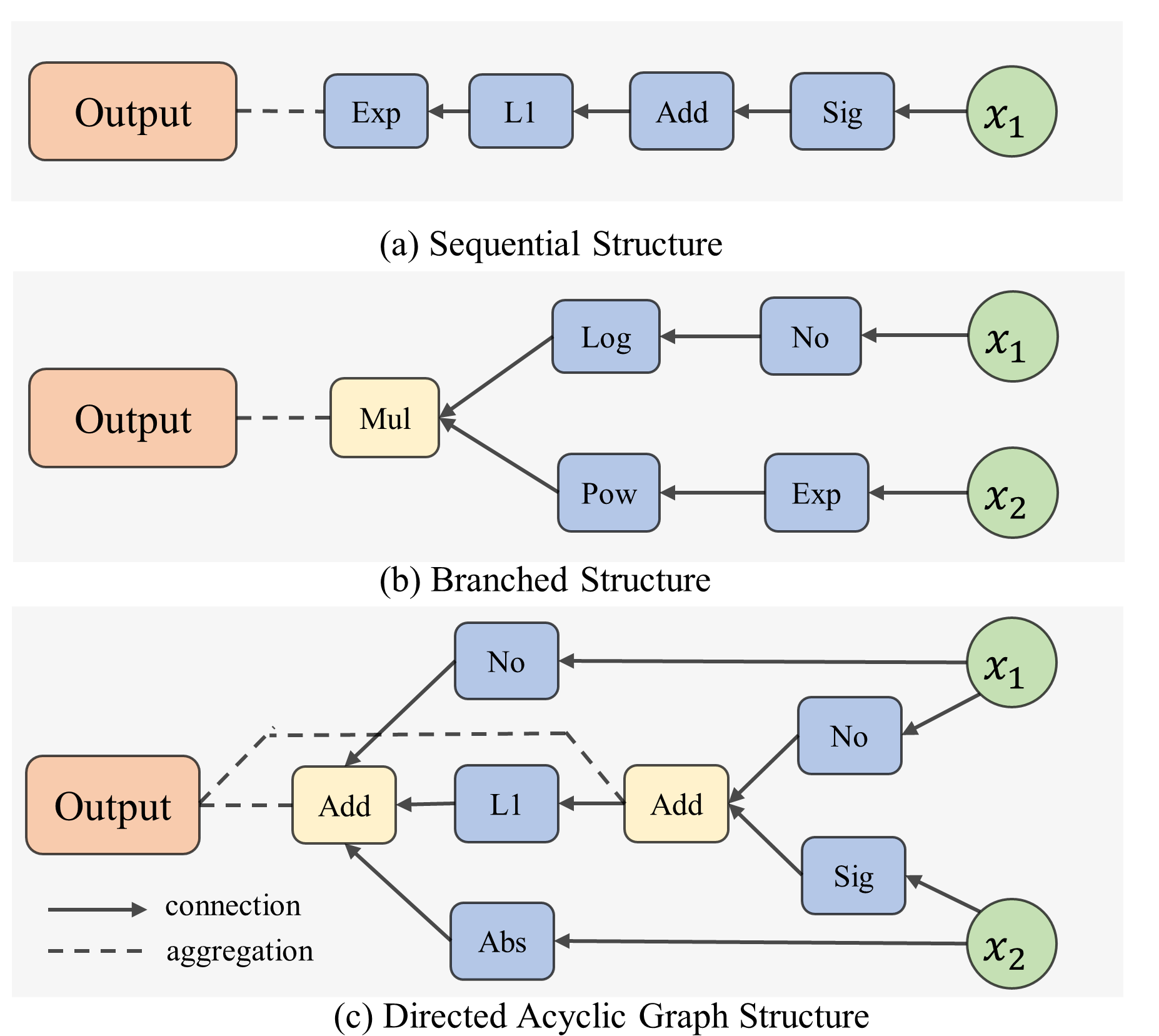}
    \end{center}
    \vspace{-0.5em}
    \caption{EMQ Search Space Structures: Three types of computation graph are shown for the EMQ search space. Dotted lines represent aggregation operations, while solid lines represent connections between nodes. (a) Sequential structure performs unary operations in a linear order. (b) Branched structure takes two branch inputs, applies various unary operations, and performs binary operations. (c) Directed Acyclic Graph (DAG) based structure takes two branch inputs and each node aggregates the statistics from all of its predecessors.}
    \label{fig:computation_graph}
\end{figure}

However, these search-based methods require an extremely large amount of computational resources and are time-consuming due to the exponential search space. One-shot methods, such as DNAS \cite{wu2018dnas} and Adabits~\cite{Jin2019AdaBitsNN}, alleviate the searching problem greatly by constructing a supernet or hypernet where each layer consists of a linear combination or parallel blocks of outputs of different bit-precisions, respectively. Nevertheless, a differentiable search for mixed-precision quantization~\cite{wu2018dnas,Habi2020HMQHF} still needs a large amount of time due to the optimization of the large hypernet.

To address the issue of bit-precision selection, heuristic criterion-based methods utilize zero-cost quantization proxies to rank the importance of layers. One approach is the Hessian-based quantization framework, which uses second-order information as the sensitivity metric. For instance, HAWQ~\cite{dong2019hawq} measures the sensitivity of each layer using the top Hessian eigenvalue and manually selects the bit-precision based on the relative sensitivity. HAWQ-V2~\cite{dong2019hawqv2} proves that the average Hessian trace is a better sensitivity metric and proposes a Pareto frontier-based method for automatic bit-precision selection. Intrinsic zero-cost proxies are also developed to handle mixed-precision quantization. QE Score~\cite{qescore} evaluates the entropy of the last output feature map without training, representing the expressiveness. OMPQ~\cite{Ma2021OMPQOM} proposes an Orthogonality Metric (ORM) that incorporates function orthogonality into neural networks and uses it to find an optimal bit configuration without any searching iterations.

The hand-crafted proxies used in previous works require expert knowledge and are often computationally inefficient~\cite{dong2019hawq,dong2019hawqv2,Ma2021OMPQOM}. These works suffer from major limitations. First, estimating the average Hessian trace using an implicit iterative approach based on the matrix-free Hutchinson algorithm~\cite{avron2011randomized} can lead to computational excesses and unstable iterative results for large-scale models. Second, the automatic bit-precision selection can only yield sub-optimal solutions, as the constraint space of the optimization problem is limited. For example, HAWQ-V2~\cite{dong2019hawqv2} considers only one constraint on memory footprint when drawing the Pareto frontier of accuracy perturbation and model size, limiting the solutions to local optima in low-dimensional spaces.
To overcome these challenges, we commence by benchmarking the existing zero-cost quantization approaches and aim to automate the process of designing zero-cost quantization proxies using techniques inspired by AutoML-zero~\cite{Real2020AutoMLZeroEM}. Our objective is to automatically search for the most effective training-free quantization proxy, capable of achieving competitive results with hand-crafted solutions.

\begin{figure}[t]
    \begin{center}
        \includegraphics[width=0.99\linewidth]{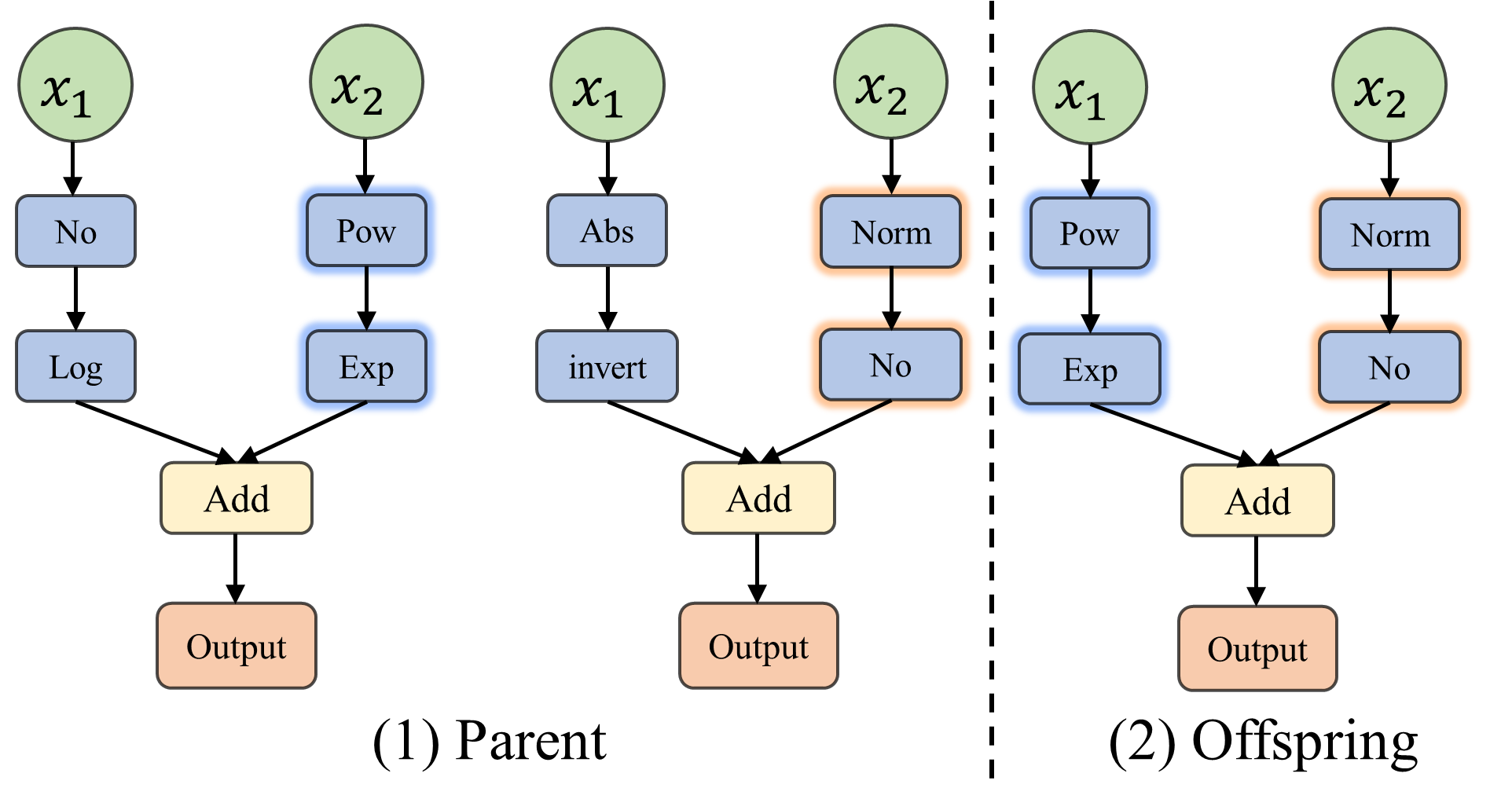}
    \end{center}
    \vspace{-0.5em}
    \caption{Crossover process for branched structures. A subtree crossover strategy is employed for the two parent structures, where a random subtree is selected from each parent and exchanged to create two new offspring structures. The glowing border indicates the selected branch.}
    \label{fig:crossover}
\end{figure}

\subsection{Zero-cost Proxies for NAS}

Recently, research has been focused on zero-shot/zero-cost neural architecture search (NAS), which estimates the performance of network architectures using zero-cost proxies based on small batches of data. Zero-shot NAS outperforms early NAS since it can estimate model performance without the need for complete training and training of super-networks in a single NAS, and without the need for forward and backward propagation of neural networks, which makes the entire process cost negligible. Zero-shot NAS is classified into two types: architecture-level and parameter-level zero-shot NAS.

(1) Architecture-level zero-shot NAS evaluates the discriminative power of different architectures through inference. For example, NWOT~\cite{mellor2020neural} found that better-performing models can better distinguish the local Jacobian values of different images and proposed an indicator based on the correlation of input Jacobian for evaluating model performance. EPE-NAS~\cite{Lopes2021EPENASEP} proposed an index based on the correlation of Jacobian with categories. ZenNAS~\cite{ZenNAS} evaluates the candidate architectures with the gradient norm of the input image as a ranking score. MAE-DET~\cite{Sun2021MAEDETRM} estimates the differential entropy of the last feature map to represent the network's expressiveness based on the maximum entropy theory~\cite{Norwich1993InformationSA}.

(2) Parameter-level zero-cost NAS aims to evaluate and prune redundant parameters from neural networks. Several indicators have been proposed for this purpose, including GradNorm~\cite{Mozer1988SkeletonizationAT}, Plain~\cite{Mozer1988SkeletonizationAT}, SNIP~\cite{Lee2018SNIPSN}, GraSP~\cite{Wang2020PickingWT}, Fisher~\cite{Theis2018FasterGP}, Synflow~\cite{Tanaka2020PruningNN} and DisWOT~\cite{dong2023diswot}. These indicators evaluate the importance of each parameter in the network and rank them based on their values.

While both types aim to alleviate the computational burden of traditional NAS, parameter-level zero-shot NAS has gained more attention due to its similarity with existing MQ proxies. Zero-cost proxies operate at the parameter level and are useful in measuring the sensitivity of each layer in a neural network. Parameter-level zero-cost proxies offer a more fine-grained approach to evaluating the performance of different network architectures, which can be used to optimize the overall performance of the system. Thus, this approach is of great value to the development of efficient and effective neural architectures. Inspired by the existing MQ proxies, we adopt the zero-cost proxies in neural architecture search to measure the sensitivity of each layer by weighting the bit-width. 

\subsection{Revisit the Zero-cost Proxies for Mixed-precision Quantization}

There are four types of input for the zero-cost quantization proxies, which are Hessian, activation, gradient, and weight. The notations are as follows: $\mathcal{L}$ denotes the loss function of a neural network, which measures the discrepancy between the predicted outputs and ground-truth labels. $\theta$ represents the parameter of a neural network, which is optimized through back-propagation with the aim of minimizing the loss function. $H$ denotes the Hessian matrix, which describes the curvature of the loss function around a particular parameter configuration. $F$ or $z$ is the activation function, which transforms the input signal into the output signal of a neuron. $i$ is used to denote the $i$-th layer of a neural network. $||\cdot||_F$ denotes the F-normalization, which normalizes a matrix by its Frobenius norm. Finally, $Tr$ denotes the trace of a matrix, which is the sum of its diagonal elements.

\textbf{(1) Hessian as Input.} Hessian Matrix represents the second-order information. Evidence~\cite{dong2019hawq} shows that the eigenvalues of the Hessian can measure the sensitivity of a specific layer of a neural network. 
The highest Hessian Spectrum is proposed in HAWQ~\cite{dong2019hawq} as shown in Equ.~\ref{eqa:hawqv1-appendix}, with which to decide the order of finetuning blocks. 

\begin{figure}[t]
    \begin{center}
        \includegraphics[width=0.69\linewidth]{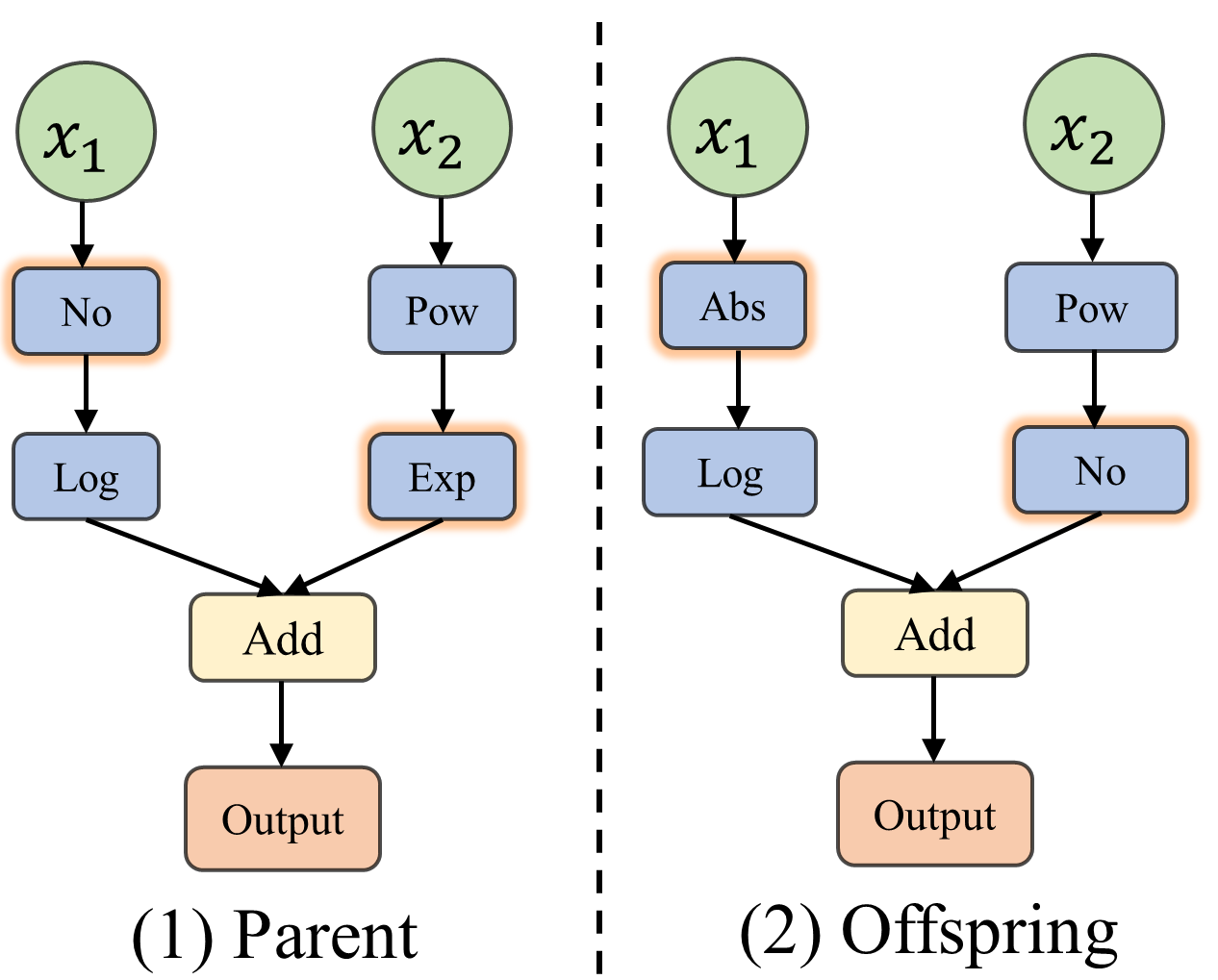}
    \end{center}
    \vspace{-0.5em}
    \caption{Mutation process for branched structures. Nodes are randomly selected and the corresponding operation would be mutated with random sampled unary operations. The glowing border indicates the selected node.}
    \label{fig:mutation}
\end{figure}

\noindent where $H$ is the Hessian matrix, $\lambda_i(H)$ denotes the $i$th eigenvalue of $H$, and $n$ is the dimension of $H$. The curly braces $\left\{ \cdot \right\}$ denote a set, and the subscript $i=1$ indicates that the set starts with the first eigenvalue, while the superscript $n$ indicates that the set ends with the $n$-th eigenvalue. 

\begin{equation}\label{eqa:hawqv1-appendix} 
\operatorname{spectrum}(H) = \max_i \{\lambda_i(H)\} 
\end{equation}

HAWQv2~\cite{dong2019hawqv2} points out the drawbacks of HAWQ for it only focuses on the top eigenvalue but ignores the rest of the Hessian spectrum. Instead, HAWQv2 adopts the average hessian trace as the MQ proxy as shown in Equ.~\ref{eqa:hwaqv2-appendix}, where $n$ is the number of Hessian matrices being averaged, $\operatorname{tr}(H_i)$ denotes the trace of the $i$th Hessian matrix, and $\frac{1}{n}\sum_{i=1}^n$ represents the average over all $n$ matrices.

\begin{equation}\label{eqa:hwaqv2-appendix}
\operatorname{trace}(H) = \frac{1}{n}\sum_{i=1}^n \operatorname{tr}(H_i)
\end{equation} 

\textbf{(2) Activation as Input.} Taking the activation as input, OMPQ~\cite{Ma2021OMPQOM} proposed the network orthogonality as the zero-cost proxy, which can correlate with the accuracy with different quantization configurations. If a neural network with $N$ layers, the activation from different layers are $\left\{F\right\}_{i}^{N}$. Then the orthogonality metric is shown in Equ.~\ref{eqa:ompq-appendix}. 

\begin{equation}\label{eqa:ompq-appendix}
\operatorname{orm}(F) = \frac{||F_j^T F_i||^2_{F}}{||F_i^T F_i||^2_{F} ||F_j^T F_j||^2_{F}}
\end{equation}

QE Score (Quantization Entropy Score)~\cite{qescore} regard neural network as an information system, propose an entropy-driven zero-cost proxy to measure the expressiveness of bit configurations. The equation is shown in Equ.~\ref{eqa:qe-appendix}, 

\begin{equation}\label{eqa:qe-appendix}
\operatorname{qescore}(F) = \sum_{l=1}^L \log\left[\frac{C_l \sigma^2 \sigma_{\text{act}}^2}{\sigma_{\text{act}}^2}\right]+\log(\sigma_{\text{act}}^2)
\end{equation}

\noindent where $C_l$ represents the product of the kernel size $K_l$ and the number of input channels $C^{l-1}$ for layer $l$. $\sigma^2$ represents the variance of the quantization value used to represent the weights. $\sigma_{\text{act}}^2$ represents the variance of the activation.  Fisher~\cite{Theis2018FasterGP} proposes a method to quantify the importance of each activation channel in a neural network, which can be used to inform channel pruning. A metric $\mathtt{fisher}$ is defined as follows: 

\begin{equation}\label{eqa:fisher-appendix}
\operatorname{fisher}(z)=\left(\frac{\partial \mathcal{L}}{\partial z} z\right)^2
\end{equation}

\noindent where $\mathcal{S}_z$ is the saliency activation $z$.

\textbf{(3) Gradient as Input.} 
Various pruning-based techniques weight the gradient with activation or weight to measure the sensitivity of a layer. SNIP \cite{Lee2018SNIPSN} computes a saliency metric at initialization using a single minibatch of data to approximate the change in loss when a specific parameter is removed. The equation is shown in ~\ref{eqa:snip-appendix}, 

\begin{equation}\label{eqa:snip-appendix}
\operatorname{snip}(\theta)=\left|\frac{\partial \mathcal{L}}{\partial \theta} \odot \theta\right|
\end{equation}

\noindent where $\mathcal{L}$ is the loss function of a neural network with parameters $\theta$, and $\odot$ is the Hadamard product. 
Synflow \cite{Tanaka2020PruningNN} proposes a modified version of the synaptic saliency scores that avoids layer collapse during parameter pruning. The synflow proxy is computed that is simply the product of all parameters in the network, so no data is needed to compute the metric. Synflow \cite{Tanaka2020PruningNN} is defined in Equ.~\ref{eqa:synflow-appendix}, 

\begin{equation}\label{eqa:synflow-appendix} 
\operatorname{synflow}(\theta)=\frac{\partial \mathcal{R}}{\partial \theta} \odot \theta
\end{equation}

\noindent where $\frac{\partial \mathcal{R}}{\partial \theta}$ denotes the gradient of the synaptic flow loss.

\textbf{(4) Weight as Input:} The weight of a neural network is another important input to compute saliency metrics. Plain \cite{Mozer1988SkeletonizationAT} proposed a method to estimate the importance of each weight in the network, where the score is determined by removing each weight and measuring the change in performance. SNIP \cite{Lee2018SNIPSN} extended this idea by computing the saliency metric at initialization using a single minibatch of data to approximate the change in loss when a specific parameter is removed. The saliency score is defined as the absolute value of the product of the gradient and the weight, as shown in Eqn.\ref{eqa:snip-appendix}. Synflow \cite{Tanaka2020PruningNN} also employs the weight as input to compute the per-parameter saliency metric, as shown in Eqn.\ref{eqa:synflow-appendix}. The synflow proxy is computed by taking the product of the gradient and weight, and it can avoid layer collapse during parameter pruning. 

\begin{figure*}[t]
    \begin{center}
        \includegraphics[width=0.9\linewidth]{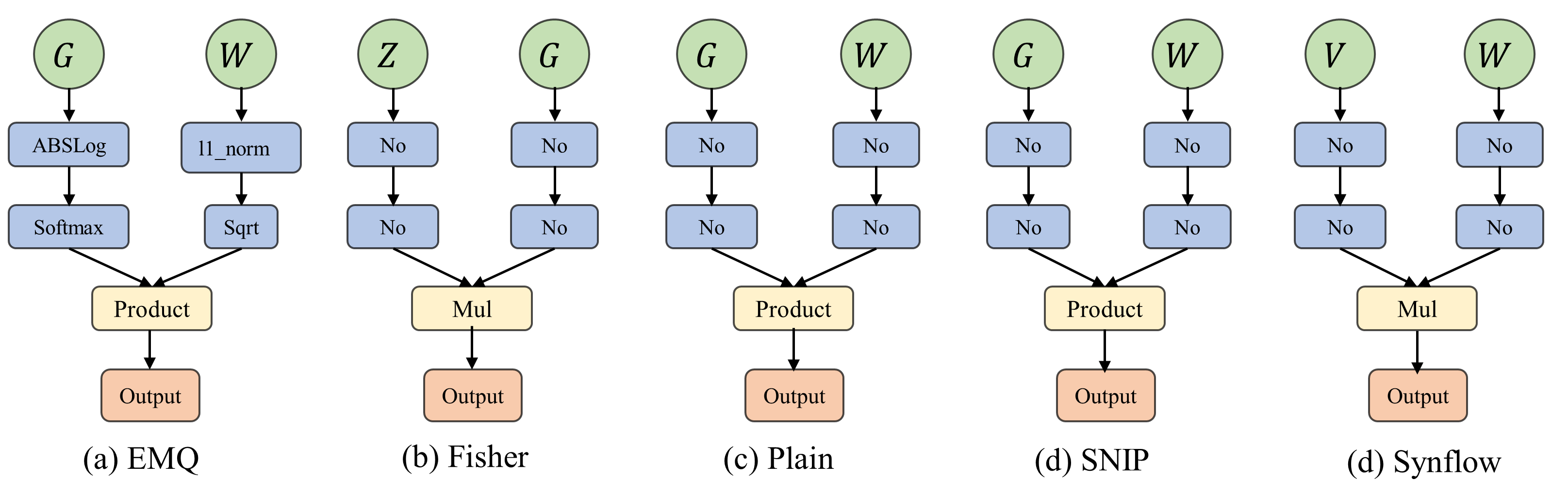}
    \end{center}
    \caption{Branched structure of different naive proxies, including the searched EMQ, Fisher~\cite{liu2021selfish}, Plain~\cite{Mozer1988SkeletonizationAT}, SNIP~\cite{Lee2018SNIPSN}, and Synflow~\cite{syflow}. The nodes represent unary operation or binary operation, and the color represents the type of operation performed in the branched structure. The searched EMQ proxy exhibits a more complex and diverse structure compared to other naive proxies, indicating its potential for achieving better compression performance.}
    \label{fig:naive_zc}
\end{figure*}

To conclude, the above studies have revealed that a variety of zero-cost quantization proxies can be constructed based upon the combination of four types of inputs (e.g. activation, gradient, weight, and hessian) and two types of operations (binary and unary).  As shown in Fig.~\ref{fig:naive_zc}, we illustrate how naive proxies, including Fisher~\cite{liu2021selfish}, Plain, SNIP~\cite{Lee2018SNIPSN}, and Synflow~\cite{syflow}, and Synflow,  represents in our branched search space, where ``G" denotes gradient, ``W" denotes weight, ``Z" denotes activation, and ``V" denotes the virtual gradient proposed in Synflow~\cite{syflow}. This motivates the exploration of a larger search space of zero-cost quantization proxies in order to identify those that demonstrate improved performance.

\section{MQ-Bench-101 Details}

For Quantization-aware training, we model the effect of quantization using simulated quantization operations, which consist of a quantizer followed by a de-quantizer.
In our implementation of post-training quantization, we utilize the ImageNet-1k dataset for training and evaluation. We set the bitwidth of activation to 8 and allow the bitwidth of weight to vary within the set $\{2, 3, 4\}$. We randomly sample 425 bit configurations and record their quantization accuracy to build the \textbf{MQ-Bench-101} (version 1.0). Moreover, we are actively training all bit configurations with a batch size of 64.  All of the experiment of MQ-Bench-101 is computed by running each bit configuration on a single GPU (NVIDIA RTX 3090).  Our chosen convolutional neural network is ResNet18, and we apply layer-wise quantization for weights. We perform weight rounding calibration with 20,000 iterations, where the temperature during calibration ranges from 20 to 2. We adopt asymmetry quantization and use mean squared error (MSE) as the scale method. The learning rate is set to 4e-4, and we use a calibration data size of 1,024. We would like to acknowledge OMPQ~\cite{Ma2021OMPQOM} for the implementation of our post-training quantization, which is based on their official repository.

\begin{table}[]
\centering
\caption{The training hyper-parameter settings of post-training quantization in MQ-Bench-101.}
\begin{tabular}{l|c}
\hline
\textbf{Setting}         & \textbf{Description}                                                                                                                                                                                                                      \\ \hline
Precision                & $\{2, 3, 4\}$                                                                                      \\
Quantization Scheme      & asymmetric                                                            \\
Calibration Data         & 1,024                                                                        \\
Weight-only Quantization      & \Checkmark \\
Activation Quantization  & 8 bit \\ 
Layer-wise Quantization & \Checkmark                                   \\ 
Infrastructure      & NVIDIA RTX 3090 \\ 
Dataset             & ImageNet-1k \\ 
Network             & ResNet18 \\ 
Learning Rate       & 4e-4 \\ 
\hline
\end{tabular}
\end{table}

To ensure the effectiveness of our implementation, we perform various experiments and evaluations. Our results show that our post-training quantization method achieves a significant reduction in memory usage without sacrificing accuracy compared to the original network. We believe our implementation can serve as a valuable reference for future researchers and practitioners working on post-training quantization for convolutional neural networks. 

\textbf{The usage of API.} We provide convenient APIs to access our MQ-Bench-101, which can be easily installed via ``pip install -e ." in our EMQ repository. The code snippet of how to use MQ-Bench-101 is given below: 

\begin{lstlisting}[language=Python]
from emq.api import EMQAPI as API 
api = API('PTQ-GT.pkl', verbose=False)
# sample random index 
rnd_idx = api.random_index()
# query by index 
acc = api.query_by_idx(rnd_idx)
# query bit_cfg by index 
bit_cfg = api.fix_bit_cfg(rnd_idx)
print(f'The index: {rnd_idx} bit_cfg: {bit_cfg}, acc: {acc:.4f}')

# sample random bit_cfg
rnd_bit_cfg = api.random_bit_cfg()
# query by bit_cfg
acc = api.query_by_cfg(rnd_bit_cfg)
# query index by bit_cfg
idx = api.get_idx_by_cfg(rnd_bit_cfg)
print(f'The index: {idx} bit_cfg: {rnd_bit_cfg}, acc: {acc:.4f}')
\end{lstlisting}
We will release the code and benchmark data file.

\section{Computation Graph Details}

\noindent \textbf{Computation Graph Structures.} The three data structures, Sequential Structure, Branched Structure, and Directed Acyclic Graph Structure, are used to represent the MQ proxies in our work. These proxies are built to search for the optimal neural architecture for a given task without requiring any training. 

(1) The sequential structure is an efficient data structure that is commonly used to represent linear computations. It is implemented as a linked list of nodes, with each node representing an operation in the computation graph. This data structure is suitable for modeling simple sequential computations. As illustrated in Fig.~\ref{fig:computation_graph} (a), we only take one type of network statistics from gradient, activation, Hessian, and weight, as input. After that, we perform 4 operations sequentially and finally perform aggregation operation, a.k.a. $\textit{to\_mean\_scalar}$, to produce the predicted score for the candidate proxy. 

(2) The branched structure is a hierarchical data structure that can represent computations with multiple branches. It is implemented as a tree, with each node representing a unary operation in the computation graph and the branches representing the binary operations through the graph. This data structure is more powerful than the sequential structure and can model more complex computations. As demonstrated in Fig.~\ref{fig:computation_graph}(b), the branched structure takes two types of network statistics as input, which can be the same or different. Then, for each branch, we sequentially perform two unary operations to transform the network statistics. After that, we perform the binary operation to take the output of the two branches as input and perform $\textit{to\_mean\_scalar}$ aggregation function to get the output. 

(3) The Directed Acyclic Graph (DAG) structure is a general-purpose data structure that can represent any computation, regardless of its complexity or structure. It is implemented as a directed graph with no directed cycles, which allows for modeling complex dependencies between nodes. This data structure provides the greatest flexibility in modeling complex computations. As shown in Fig.~\ref{fig:computation_graph}(c), we take the same input settings as the branched structure. For each node in the DAG, it would aggregate the information from all of its predecessors and here we adopt binary operation to perform the middle aggregation operation. The Fig.~\ref{fig:computation_graph} illustrates the computation graph when there are just two middle nodes. We adopt three middle nodes as the default setting, which is more complex than the current computation graph. Finally, we aggregate all of the statistics from the middle nodes to produce the final output by $\mathit{to\_mean\_scalar}$.

All three data structures are utilized in constructing the MQ proxies for searching the optimal neural architecture in our work. Nonetheless, considering the superior trade-off between validity and expressive capability, we primarily adopt the branched structure in this paper.

\noindent \textbf{CrossOver for Structures.} To perform crossover for the three computation graph structures, sequential structure, branched structure, and Directed Acyclic Graph structure, we use different strategies. Here we only depict how the branched structure performs crossover in Fig.~\ref{fig:crossover}. The probability of performing crossover is set to 50\%.

(1) For sequential structure, we adopt a simple crossover strategy. Given two parent structures, we randomly select a crossover point and exchange the remaining nodes to create two new offspring structures.

(2) For the branched structure, we use a branch-switching crossover strategy. Given two parent structures, we randomly select a branch from each parent and exchange them to create two new offspring structures. This strategy allows for exchanging complex branches between parent structures and potentially producing more diverse offspring. As depicted in Fig.~\ref{fig:crossover}, the two branches of offspring are selected from one of the branches of the parent structure. 

(3) For the Directed Acyclic Graph structure, we adopt a random graph-based crossover strategy. Given two parent structures, we randomly select a subset of nodes from each parent and exchange them to create two new offspring structures. This strategy allows for exchanging nodes with different levels of connectivity and potentially producing more diverse offspring.

\noindent \textbf{Mutation for Computation Graph Structures.} Mutation is an important operation in the evolutionary algorithm. For the Sequential and Branched structures, we perform mutation by modifying a randomly selected node, which can be either a layer or a connection between layers. Specifically, we randomly select a node and replace it with a new one sampled from the search space with the probability of 50\%, as depicted in Fig.~\ref{fig:mutation}. In the Directed Acyclic Graph structure, we perform mutation by modifying the connections between nodes. We randomly select a node and add or delete its incoming or outgoing edges, or we change the type of an existing edge. This type of mutation allows for more complex modifications to the graph structure and enables the model to explore a larger search space.

\section{Additional Details in Evolving Algorithm}

\subsection{Routine}

In the context of mixed-precision quantization, proxies play a crucial role in efficiently exploring the search space of possible bit configurations for quantization. To evaluate the quality of different configurations, there are two main routines in the MQ framework: scoring the bit configurations as a whole and evaluating the layer-wise sensitivity separately.

Scoring the bit configurations involves computing a single scalar score for a given quantization configuration, which reflects its overall performance. This routine can be done efficiently without the need for training and evaluation since it only requires analyzing the model's structure and computing the gradient of each layer's output with respect to its input.

On the other hand, evaluating layer-wise sensitivity involves measuring the sensitivity of each layer in a model to quantization. This routine can be computationally expensive since it requires training and evaluating a separate set of quantized models for each configuration, which can be time-consuming for large models.

In this paper, we focus on tackling the former routine of scoring the bit configurations. We propose a novel approach that leverages a proxy to estimate the performance of a given quantization configuration without the need for training and evaluating separate models. Our approach relies on learning a function that maps a quantization configuration to a corresponding performance score. To train this function, we use a set of precomputed scores for a subset of quantization configurations and a gradient-based optimization method to fit the function to the available data. Our proposed method achieves competitive results on benchmark datasets while significantly reducing the computational cost of searching for optimal quantization configurations.

\subsection{Weisfeiler-lehman test}

The Weisfeiler-Lehman (WL) test is a powerful method for graph isomorphism testing, which can also be used to perform the de-isomorphism process for the computation graph structures.

To apply the WL test, we first initialize each node in the graph with a unique label. Then, for each iteration, we follow these steps: For each node, we collect the labels of its neighbors and sort them. We concatenate the sorted neighbor labels and the node's own label into a new string. We use a hash function to map the new string to a new label for the node. We update the label of each node with its new label. We repeat these steps for a fixed number of iterations, denoted by $h$. After $h$ iterations, we obtain a new set of labels for all nodes in the graph, which can be used to perform the de-isomorphism process.

\begin{table}[t]
\centering
\small 
\caption{The affection of different initialization seeds on the overall Spearman rank correlation. The rank consistency results are conducted on MQ-Bench-101.}
\resizebox{\linewidth}{!}{
\begin{tabular}{c|cccc|cc}
\hline
\multirow{2}{*}{Proxy} & \multicolumn{4}{c|}{Seed}         & \multirow{2}{*}{MEAN} & \multirow{2}{*}{STD} \\ \cline{2-5}
                       & 0      & 1      & 2      & 3      &                       &                      \\ \hline
BParam                 & 0.3353 & 0.6031 & 0.7082 & 0.5564 & 0.5508                & 0.1360               \\
SNIP                  & 0.3334 & 0.4603 & 0.2605 & 0.4850 & 0.3848                & 0.0920               \\
Synflow               & 0.2841 & 0.3398 & 0.2991 & 0.3398 & 0.3157                & 0.0247               \\
HAWQ                 & 0.5821 & 0.5316 & 0.7233 & 0.5821 & 0.6048                & 0.0715               \\
HAWQ-V2                 & 0.7813 & 0.8243 & 0.6943 & 0.6903 & 0.7475                & 0.0573               \\
OMPQ                   & 0.3295 & 0.3406 & 0.3141 & 0.2585 & 0.3107                & 0.0316               \\
QE                     & 0.3280 & 0.3519 & 0.3185 & 0.4616 & 0.3650                & 0.0571               \\
EMQ                    & 0.7132 & 0.7841 & 0.8712 & 0.7998 & 0.7921                & 0.0561               \\ \hline
\end{tabular}}\label{tab:seed_ablation}
\end{table}

\begin{table*}[t]
    \centering
    \caption{The unary operations and binary operations in the search space. ``UOP" denotes the unary operations, and ``BOP" denotes the binary operations. The type of input and output can be scalar or matrix. ``no\_op" denotes that we do not perform any operation, and allows for the sparsity of the computation graph. Not all operations below are available or mathematically sound. When there is an illegal operation, we adopt a try-catch mechanism to detect the invalidity and avoid the interruption of the search process.}
    \resizebox{0.99\linewidth}{!}{
    \begin{tabular}{l|llll}
        \toprule[1.1pt]
        OP ID & OP Name                                  & Input Args        & Output Args   & Description                                         \\    \midrule[0.8pt]
        UOP00   & \textcode{no\_op}                        & --                & --            & --                                                  \\ \midrule[0.8pt]
        UOP01 & \textcode{element\_wise\_abs}            & $a$ / scalar,matrix     & $b $ / scalar,matrix & $x_b=\left|x_a\right|$                              \\
        UOP02 & \textcode{element\_wise\_tanh}            & $a$ / scalar,matrix     & $b $ / scalar,matrix & $x_b=\tanh(x_a)$                                          \\
        UOP03 & \textcode{element\_wise\_pow}            & $a$ / scalar,matrix & $b$ / scalar,matrix & $x_b=x_a^2$                                       \\
        UOP04 & \textcode{element\_wise\_exp}            & $a$ / scalar,matrix     & $b $ / scalar,matrix & $x_b=e^{x_a}$                                       \\
        UOP05 & \textcode{element\_wise\_log}            & $a$ / scalar,matrix     & $b $ / scalar,matrix & $x_b=\ln x_a$                                       \\
        UOP06 & \textcode{element\_wise\_relu}           & $a$ / scalar,matrix     & $b $ / scalar,matrix & $x_b=\max(0,x_a)$                                   \\
        UOP07 & \textcode{element\_wise\_leaky\_relu}           & $a$ / scalar,matrix     & $b $ / scalar,matrix & $x_b=\max(0.1x_a,x_a)$                                   \\
        UOP08 & \textcode{element\_wise\_swish}           & $a$ / scalar,matrix     & $b $ / scalar,matrix & $x_b=x_a\times \text{sigmoid}(x_a)$                                   \\
        UOP09 & \textcode{element\_wise\_mish}           & $a$ / scalar,matrix     & $b $ / scalar,matrix & $x_b=x_a\times \text{tanh}(\ln{1+\text{exp}(x_a)})$                                   \\
        UOP10 & \textcode{element\_wise\_invert}         & $a$ / scalar,matrix     & $b $ / scalar,matrix & $x_b=1/x_a$                                         \\
        UOP11 & \textcode{element\_wise\_normalized\_sum} & $a$ / scalar,matrix & $b$ / scalar,matrix & $x_b=\frac{\sum x_a}{\text{numel}(x_a)+\epsilon}$                     \\
        UOP12 & \textcode{normalize}                     & $a$ / scalar,matrix     & $b $ / scalar,matrix & $x_b=\frac{x_a - \text{mean}(x_a)}{\text{std}(x_a)}$                    \\
        UOP13 & \textcode{sigmoid}                       & $a$ / scalar,matrix     & $b $ / scalar,matrix & $x_b=\frac{1}{1+e^{-x_a}}$                          \\
        UOP14 & \textcode{logsoftmax}                    & $a$ / scalar,matrix     & $b $ / scalar,matrix & $x_b=\ln\frac{e^{x_a}}{\sum_{i=1}^n e^{s_i}}$       \\
        UOP15 & \textcode{softmax}                       & $a$ / scalar,matrix     & $b $ / scalar,matrix & $x_b=\frac{e^{x_a}}{\sum_{i=1}^n e^{s_i}}$          \\
        UOP16 & \textcode{element\_wise\_sqrt}           & $a$ / scalar,matrix     & $b $ / scalar,matrix & $x_b=\sqrt{x_a}$                                    \\
        UOP17 & \textcode{element\_wise\_revert}         & $a$ / scalar,matrix     & $b $ / scalar,matrix & $x_b=-x_a$                                         \\
        UOP18 & \textcode{frobenius\_norm}                & $a$ / scalar,matrix     & $b $ / scalar,matrix & $x_b=\sqrt{\sum_{i=1}^n s_i^2}$                     \\
        UOP19 & \textcode{element\_wise\_abslog}         & $a$ / scalar,matrix     & $b $ / scalar,matrix & $x_b=\left|\ln x_a\right|$                          \\
        UOP20 & \textcode{l1\_norm}                      & $a$ / scalar,matrix     & $b $ / scalar,matrix & $x_b=\frac{\sum_{i=1}^n \left|s_i\right|}{\text{numel}(x_a)}$                 \\
        UOP21 & \textcode{min\_max\_normalize}           & $a$ / scalar,matrix     & $b $ / scalar,matrix & $x_b=\frac{x_a-\min(x_a)}{\max(x_a)-\min(x_a)}$     \\
        UOP22 & \textcode{to\_mean\_scalar}              & $a$ / scalar,matrix     & $b $ / scalar & $x_b=\frac{x_a}{n}$                                 \\
        UOP23 & \textcode{to\_std\_scalar}               & $a$ / scalar,matrix     & $b $ / scalar & $x_b=\sqrt{\frac{\sum_{i=1}^n (s_i-\bar{s})^2}{n}}$ \\ \midrule[0.8pt]

        BOP01 & \textcode{element\_wise\_sum}            & $a$,$b$ / scalar,matrixs & $c $ / scalar,matrix & $x_c=x_a+x_b$                                       \\
        BOP02 & \textcode{element\_wise\_difference}     & $a$,$b$ / scalar,matrixs & $c $ / scalar,matrix & $x_c=x_a-x_b$                                       \\
        BOP03 & \textcode{element\_wise\_product}        & $a$,$b$ / scalar,matrixs & $c $ / scalar,matrix & $x_c=x_a\times x_b$                                       \\
        BOP04 & \textcode{matrix\_multiplication}        & $a$,$b$ / scalar,matrixs & $c $ / scalar,matrix & $x_c=x_a @x_b$                                     \\ \bottomrule[1.1pt]
    \end{tabular}}
\end{table*}

\subsection{Conflict Awareness}

Conflict awareness is a crucial aspect of optimizing search processes. In many cases, different operations that are part of the search space can conflict with each other, leading to unexpected or unstable behavior. In this section, we explore some common examples of conflicting operations and their potential impact on the search process. The potential conflict operation pairs are summarized as follows:

\begin{itemize}
    \item ``log" and ``exp": These operations are inverses of each other, so applying them in succession may effectively cancel each other out. This can lead to numerical instability, especially when dealing with small or large values. 
    \item ``normalize" and ``min\_max\_normalize": Both of these operations involve scaling the input data to lie within a certain range. However, they use different scaling strategies, which can cause conflicts when applied in succession. For example, normalizing data to have zero mean and unit variance (as in ``normalize") may undo the effects of min-max normalization, which scales the data to lie within a specified range.
    \item ``relu" and ``sigmoid": These activation functions have different properties and are often used for different purposes. ReLU is commonly used for its simplicity and efficiency in deep neural networks, while sigmoid is often used in binary classification tasks. However, applying these functions in succession can lead to non-monotonic behavior, which can cause optimization problems.
    \item ``log" and ``softmax": Both of these operations involve taking the logarithm of the input data. However, the softmax function also involves exponentiation and normalization, which can lead to numerical instability when combined with the logarithm function.
    \item ``pow" and ``sqrt": pow raises a number to a power, while sqrt calculates the square root. Using them together may lead to unexpected results or loss of precision.
    \item ``sigmoid" and ``softmax": these operations are commonly used in neural networks, but applying them together may lead to overfitting or unstable behavior.
    \item ``frobenius\_norm" and ``revert": frobenius\_norm calculates the Frobenius norm of a matrix (i.e., the square root of the sum of the squared values). Applying revert to a matrix will negate all its values, including the norm.
     \item ``invert" and ``revert" operations are also in conflict with themselves. The invert operation involves dividing 1 by the tensor, which can cause numerical instability when the tensor contains values close to 0. The revert operation involves subtracting the tensor from 1, which can also cause numerical instability when the tensor contains values close to 1. 
     \item ``abs" and ``relu": While these operations are not mathematically inverse, they have similar effects on the input data. Using them together in the same search space may lead to redundant or contradictory combinations. 
     \item ``to\_mean\_scalar" and ``to\_sum\_scalar": In MQ proxy discovery, we do not care for the absolute value of the proxy score but the relative value. To aggregation operations, computing the mean of the value and the sum of it do not influence the ranking ability or correlation of a MQ proxy. 
     \item others: In general, there are conflicts between activation functions, such as ``relu",  ``leaky\_relu", ``swish" and ``mish",  and two consecutive activation functions do not need to appear in a computational graph.
\end{itemize}

Generally, conflict awareness plays a non-trivial role in optimizing search processes. It helps to mitigate numerical instability, improve performance, and ensure the efficiency and effectiveness of the search process. Our analysis of common examples of conflicting operations emphasizes the need to consider the relationships between different operations and their potential impact on the overall search process. By identifying potentially conflicting operation pairs, such as those we have discussed, the EMQ search process can avoid generating invalid combinations of operations and instead focus on discovering high-quality configurations that are both diverse and effective. Ultimately, conflict awareness is a critical component of any search process that aims to produce accurate and reliable results.

\subsection{Naive Invalid Check}

In the search for an optimal mixed-precision quantization scheme, the validity of the generated proxies is crucial. The Naive Invalid Check technique is a simple yet effective way to reduce the number of invalid proxies generated during the search process. This technique involves checking if the estimated score of a proxy belongs to a set of invalid scores, which includes $\{-1, 1, 0, \textit{NaN}, \textit{and Inf}\}$. These scores indicate that the proxy is indistinguishable and unreliable, and should be rejected at an early stage.

An estimated score of a MQ proxy in EMQ search space can be invalid due to several reasons. For example, a score of -1 arise from a shape mismatch issue or a user-defined exception, while a score of 1 may indicate a numerical insensitivity. A score of 0 may indicate a numerical instability or the result of an invalid operation. NaN (Not a Number) may arise due to a variety of reasons such as division by zero, square root of a negative number, or logarithm of a non-positive number. Finally, a score of infinity may arise from an overflow in arithmetic operations or the result of invalid mathematical operations. By rejecting these invalid proxies early, the search space is reduced, and the optimization process becomes more efficient. Furthermore, the rejection of invalid proxies reduces the computational cost of evaluating the fitness of the generated proxies, which can be quite expensive. Despite its simplicity, the Naive Invalid Check technique has been shown to be effective in identifying invalid proxies, as the set of invalid scores used in the technique covers a wide range of possible invalid proxy configurations.

\subsection{Operation Sampling Prioritization}
To mitigate the large number of invalid candidates that result from random operation sampling during the search for MQ proxies, we propose Operation Sampling Prioritization (OSP), which assigns different probabilities to various operations. Specifically, we assign a higher probability to the no op operation for unary operations to promote sparsity in the search space and prevent an excess of operations. For binary operations, we assign a higher probability to the element-wise add operation, as it is the most common operation and unlikely to cause shape-mismatch problems. We also assign probabilities to other binary operations based on their likelihood to cause shape-mismatch problems. Concretely,

\begin{itemize}
\item For unary operations: We assign a probability of 0.2 to the "no\_op" operation to promote sparsity in the search space and prevent excessive operations. The remaining operations are assigned an equal probability of 0.1.
\item For binary operations: We assign a probability of 0.6 to the "sum" operation, which is the most common and least likely to cause shape-mismatch problems. We assign a probability of 0.3 to the "subtract" operation. The "product" and "matrix\_multiplication" operations, which are more likely to cause shape-mismatch problems, are assigned a lower probability of 0.05.
\end{itemize}

\subsection{Iterations and Population} 
\begin{figure}
    \centering
    \includegraphics[width=0.89\linewidth]{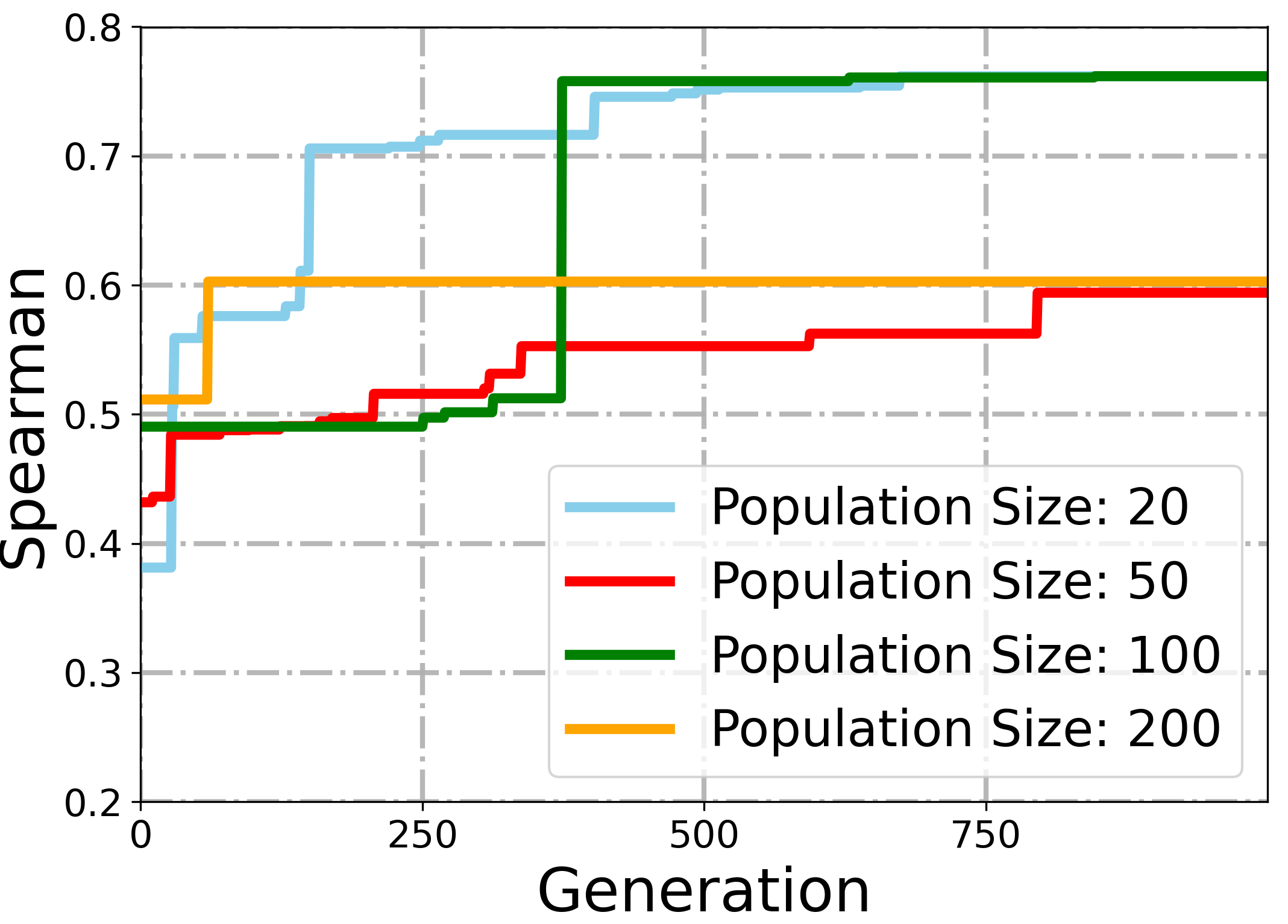}
    \caption{Effect of population size on initialization Spearman correlation. As the population size increases, the initialization Spearman correlation improves.}
    \label{fig:popusize}
\end{figure}

In this section, we discuss how we determined the appropriate iteration and population size for the evolutionary algorithm. Both iteration and population size are important parameters that affect the performance of evolutionary algorithms. To determine the iteration, we used a stopping criterion that takes the best Spearman rank coefficient of human-designed MQ proxies~\cite{syflow,Lee2018SNIPSN,Ma2021OMPQOM, qescore} as the desired level of correlation. For population size, we found that it mainly affects initialization performance. Thanks to the proposed diversity-prompting selection mechanism, we can maintain population diversity with a small population size of 20. As shown in Fig.~\ref{fig:popusize}, the Spearman coefficient for the initialization generation with a population size of 200 is over 0.5, surpassing the one with a population size of 20 by a large margin (10\%). However, the population size only influences initialization performance and does not affect the final performance.

\subsection{Data Splitting in Search Process}
To search for zero cost proxies for quantization, we divided the MQ-Bench-101 dataset into a validation set (70\%) and a test set (30\%). In the search phase, we randomly selected 50 bit-widths from the validation set. We then evaluated the performance of our proxy using the test set, ensuring an unbiased assessment of the ranking consistency. Furthermore, we made sure that there was no overlap between the validation and test sets to guarantee the fairness and impartiality of all experiments and comparisons. Unlike training-based methods, our proxy discovery process is training-free, which means that no training dataset was used in the search process.

\section{More Ablation Study}
In this section, we will perform ablation studies to analyze the sensitivity of the searched EMQ proxy to different settings. This will help us understand the impact of these parameters on the performance of the search process and identify the optimal settings for achieving high-quality results.

\subsection{Sensitivity Analysis of Batch Size} 
\begin{figure}[t]
    \centering
    \includegraphics[width=0.99\linewidth]{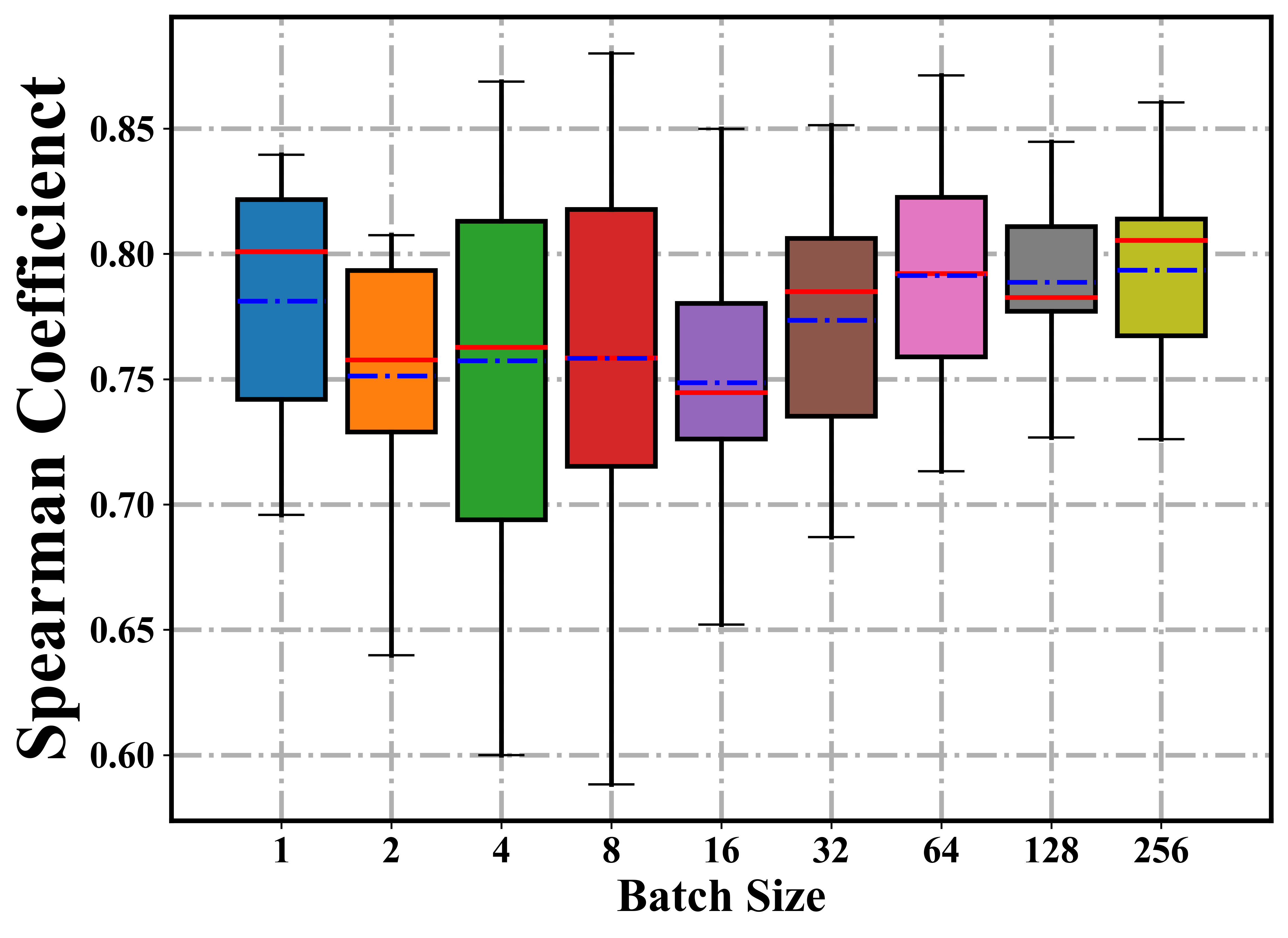}
    \caption{Effect on Spearman rank coefficient of the searched EMQ with respect to batch size. The experiments are done on the MQ-Bench-101 over 7 seeds for 50 bit configurations}
    \label{fig:emq_box}
\end{figure}

We present the sensitivity analysis of the searched EMQ proxy to batch size. We observed that the searched EMQ proxy is completely data-free, and when different batch sizes are used with the same seed, the Spearman rank correlation remains the same. This is intuitive because the searched EMQ takes the gradient of the Synaptic flow loss~\cite{syflow} and the weight, both of which are unrelated to the input. We also noted that when we set "shuffle=False" in the dataloader, the Spearman's rank correlation remains unaffected by the batch size. However, shuffling the mini-batch of data during evaluation can be influenced by the seed, which in turn affects the mini-batch of data. The Fig.~\ref{fig:emq_box} investigates the impact of batch size on the Spearman correlation of the searched EMQ proxy. The results reveal that increasing the batch size leads to an improvement in the average Spearman correlation. Moreover, the variance of the Spearman correlation over seven seeds decreases as the batch size increases. Notably, when the batch size is extremely small, the Spearman correlation exhibits a surprisingly good performance. Based on the above observation, we select 64 as the batch size to strike a balance between computational complexity and correlation performance. Specifically, selecting a batch size that is too small may improve correlation performance but will increase computational overhead due to frequent parameter updates, while choosing a batch size that is too large can lead to slower convergence and worse correlation performance. However, under extreme computational constraints, a batch size of 1 can be selected.

\subsection{Sensitivity Analysis of Seeds} To assess the influence of the random seed and batch on the searched MQ proxies, we conduct experiments using different seeds and batches of data. This investigation aids in comprehending the extent to which variations in search outcomes can be ascribed to the random seed and batch data. By performing a sensitivity analysis of the seed, we can ensure the robustness of the searched MQ proxy and minimize its susceptibility to the influence of a specific random seed. The seed mainly influences the candidate bit configuration chosen during the evaluating the EMQ proxy. When the seed is fixed, the sampled bit configuration is also fixed, resulting in the same outcomes. As indicated in Tab.~\ref{tab:seed_ablation}, we conduct each experiment four times using different seeds and compute the mean and standard deviation of the Spearman rank correlation. Our searched EMQ proxy exhibit the best correlation on MQ-Bench-101, with a similar variance as other unsophisticated proxies. Notably, we observe that when scoring bit configurations as a whole, HAWQ~\cite{dong2019hawq} and HAWQ-V2~\cite{dong2019hawqv2} also achieve competent performance when compared with their counterparts.

\subsection{Performance with other metrics}

Tab.~\ref{tab:more_rank} presents the results of the ranking consistency analysis using Kendall's Tau and Pearson correlation coefficients. The experiments were run five times with different seeds, and the ranking correlation was computed based on 50 bit configurations. The rankings were compared between the different runs to evaluate the consistency of the results. 

Kendall's Tau and Pearson are two widely used correlation coefficients in data analysis. Kendall's Tau is a non-parametric measure of the association between two variables, which means that it does not assume any particular distribution of the data. It measures the similarity of the rankings between two variables, and it ranges between -1 (perfect negative correlation) and 1 (perfect positive correlation). Pearson correlation coefficient, on the other hand, assumes a linear relationship between two variables and measures the strength of this relationship. It ranges between -1 (perfect negative correlation) and 1 (perfect positive correlation).

The results of Kendall's Tau and Pearson correlation coefficients furnish us with a more extensive comprehension of the effectiveness of the hand-crafted MQ proxy and our designed EMQ proxy. It is noteworthy that our EMQ proxy demonstrates the highest correlation in both Kendall's Tau and Pearson. It is also notable that the HAWQ~\cite{dong2019hawq} and HAWQv2~\cite{dong2019hawqv2} exhibit proficient performance.

\begin{table}[]
\centering
\caption{The ranking consistency with Kendall's Tau and Pearson. Each experiments run 5 times and the ranking correlation is computed based on 50 bit configurations.}
\resizebox{0.85\linewidth}{!}{
\begin{tabular}{l|cc|cc}
\hline
\textbf{} & \multicolumn{2}{c}{Kendall Tau} & \multicolumn{2}{c}{Pearson} \\ \hline
          & MEAN           & STD            & MEAN         & STD          \\ \hline
QE~\cite{qescore}        & 0.2831         & 0.0493         & 0.4131       & 0.0808       \\
OMPQ~\cite{Ma2021OMPQOM}      & 0.1766         & 0.0249         & 0.1823       & 0.0312       \\
HAWQv2~\cite{dong2019hawqv2}    & 0.5550         & 0.0478         & 0.7213       & 0.0486       \\
HAWQ~\cite{dong2019hawq}    & 0.4722         & 0.0006         & 0.6795       & 0.0028       \\
Synflow~\cite{syflow}   & 0.2356         & 0.0768         & 0.3633       & 0.1130       \\
SNIP~\cite{Lee2018SNIPSN}      & 0.2724         & 0.0175         & 0.2313       & 0.0227       \\
Bparam    & 0.3133         & 0.0893         & 0.4763       & 0.1318       \\
EMQ(Ours)       & 0.6030         & 0.0373         & 0.8084       & 0.0315       \\ \hline
\end{tabular}}\label{tab:more_rank}
\end{table}

\section{Primitive Operations}

The primitive operations used in the search space of EMQ can be classified into two categories: unary and binary operations.
\begin{itemize}
    \item The unary operations available in the search space include ``log", ``abslog", ``abs", ``pow", ``exp", ``normalize", ``relu", ``swish", ``mish", ``leaky\_relu", ``tanh", ``invert", ``frobenius\_norm", ``normalized\_sum", ``l1\_norm", ``softmax", ``sigmoid", ``logsoftmax", ``sqrt", ``revert", ``min\_max\_normalize", ``to\_mean\_scalar", ``to\_std\_scalar", and ``no\_op." These operations can be applied to a single input tensor, which may have any number of dimensions.
    \item The binary operations available in the search space are ``sum", ``subtract", ``multiply", and ``dot." These operations can be applied to two input tensors, which may have any number of dimensions, as long as they are compatible for the specified operation.
\end{itemize}

The possible types of statistics in the computation graph can be scalar, vector o r graph. However, we observe that the computation between the matrix and vector is always mismatched and can not function well. Most of the middle statistics in the computation graph are matrix type, which would cause severe shape mismatch problems and decrease the search process of the EMQ. Empirically, we the operations that can produce the type of vector, for example, ``heaviside", ``dot", ``std", etc. 

\section{Joint Architecture and Bit-width Search} 

We allows for joint architecture and bit-width search for both weight and activation, following the QE~\cite{qescore} repository. This means that we can perform simultaneous optimization of the architecture and the bit-widths of both weights and activations. We implemented this approach and evaluated its performance using a larger search space. The results of our experiments are shown in Table \ref{reb:tab:1}, which demonstrates the effectiveness of our approach, EMQ, when applied to this expanded search space. Notably, our approach achieves significant improvements compared to existing methods and demonstrates the potential for further optimization. These findings suggest that our joint architecture and bit-width search approach can be a promising direction for efficient neural network design and optimization.

\begin{table}[]
\caption{Joint architecture and bit-width search for activation and weight.}\label{reb:tab:1}
\resizebox{0.48\textwidth}{!}{
\begin{tabular}{ccr|ccr}
\hline
\multicolumn{1}{l}{} & Acc(\%)   & Size(MB) & \multicolumn{1}{l}{} & Acc(\%)   & Size(MB) \\ \hline
ResNet18             & \textbf{69.76} & 11.67   & GhostNetx0.5         & 66.2  & 2.6     \\
DY-ResNet-10         & 67.70  & 1.82    & HVT-Ti-1             & 69.64 & 5.74    \\
SimpleNetv1-small    & 69.11 & 1.5   & BRECQ(MBv2)          & 68.99 & 1.3        \\
MUXNet-xs            & 66.70 & 1.8     & EMQ(Ours)      & \textbf{69.54} & \textbf{0.88}    \\ \hline
\end{tabular}
}
\end{table}

{\small
\bibliographystyle{ieee_fullname}
\bibliography{egbib}

\begin{thebibliography}{10}\itemsep=-1pt

\bibitem{avron2011randomized}
Haim Avron and Sivan Toledo.
\newblock Randomized algorithms for estimating the trace of an implicit
  symmetric positive semi-definite matrix.
\newblock {\em Journal of the ACM (JACM)}, 58(2):1--34, 2011.

\bibitem{cai2020zeroq}
Yaohui Cai, Zhewei Yao, Zhen Dong, Amir Gholami, Michael~W Mahoney, and Kurt
  Keutzer.
\newblock Zeroq: A novel zero shot quantization framework.
\newblock In {\em Proceedings of the IEEE/CVF Conference on Computer Vision and
  Pattern Recognition}, pages 13169--13178, 2020.

\bibitem{Cai2020RethinkingDS}
Zhaowei Cai and Nuno Vasconcelos.
\newblock Rethinking differentiable search for mixed-precision neural networks.
\newblock {\em 2020 IEEE/CVF Conference on Computer Vision and Pattern
  Recognition (CVPR)}, pages 2346--2355, 2020.

\bibitem{chen2019pdarts}
Xin Chen, Lingxi Xie, Jun Wu, and Qi Tian.
\newblock Progressive differentiable architecture search: Bridging the depth
  gap between search and evaluation.
\newblock {\em CoRR}, abs/1904.12760, 2019.

\bibitem{Chin2020OneWB}
Ting-Wu Chin, Pierce I-Jen Chuang, Vikas Chandra, and Diana Marculescu.
\newblock One weight bitwidth to rule them all.
\newblock {\em ArXiv}, abs/2008.09916, 2020.

\bibitem{chin2020one}
Ting-Wu Chin, I Pierce, Jen Chuang, Vikas Chandra, and Diana Marculescu.
\newblock One weight bitwidth to rule them all.
\newblock In {\em European Conference on Computer Vision}, pages 85--103.
  Springer, 2020.

\bibitem{choi2018pact}
Jungwook Choi, Zhuo Wang, Swagath Venkataramani, Pierce I-Jen Chuang,
  Vijayalakshmi Srinivasan, and Kailash Gopalakrishnan.
\newblock Pact: Parameterized clipping activation for quantized neural
  networks.
\newblock {\em arXiv preprint arXiv:1805.06085}, 2018.

\bibitem{dong2023diswot}
Peijie Dong, Lujun Li, and Zimian Wei.
\newblock Diswot: Student architecture search for distillation without
  training.
\newblock In {\em CVPR}, 2023.

\bibitem{dong2023rd}
Peijie Dong, Xin Niu, Lujun Li, Zhiliang Tian, Xiaodong Wang, Zimian Wei,
  Hengyue Pan, and Dongsheng Li.
\newblock Rd-nas: Enhancing one-shot supernet ranking ability via ranking
  distillation from zero-cost proxies.
\newblock {\em arXiv preprint arXiv:2301.09850}, 2023.

\bibitem{dong2019hawqv2}
Zhen Dong, Zhewei Yao, Yaohui Cai, Daiyaan Arfeen, Amir Gholami, Michael~W
  Mahoney, and Kurt Keutzer.
\newblock Hawq-v2: Hessian aware trace-weighted quantization of neural
  networks.
\newblock {\em arXiv preprint arXiv:1911.03852}, 2019.

\bibitem{dong2019hawq}
Zhen Dong, Zhewei Yao, Amir Gholami, Michael~W Mahoney, and Kurt Keutzer.
\newblock Hawq: Hessian aware quantization of neural networks with
  mixed-precision.
\newblock In {\em Proceedings of the IEEE/CVF International Conference on
  Computer Vision}, pages 293--302, 2019.

\bibitem{Elthakeb2020ReLeQA}
Ahmed~T. Elthakeb, Prannoy Pilligundla, FatemehSadat Mireshghallah, Amir
  Yazdanbakhsh, and Hadi Esmaeilzadeh.
\newblock Releq : A reinforcement learning approach for automatic deep
  quantization of neural networks.
\newblock {\em IEEE Micro}, 40:37--45, 2020.

\bibitem{LSQ}
Steven~K. Esser, Jeffrey~L. McKinstry, Deepika Bablani, Rathinakumar Appuswamy,
  and Dharmendra~S. Modha.
\newblock Learned step size quantization.
\newblock {\em ArXiv}, abs/1902.08153, 2019.

\bibitem{gou2020knowledge}
Jianping Gou, Baosheng Yu, Stephen~John Maybank, and Dacheng Tao.
\newblock Knowledge distillation: A survey, 2020.

\bibitem{Gu2022AutoLossGMSSG}
Hongyang Gu, Jianmin Li, Guang zhi Fu, Chifong Wong, Xinghao Chen, and Jun Zhu.
\newblock Autoloss-gms: Searching generalized margin-based softmax loss
  function for person re-identification.
\newblock {\em 2022 IEEE/CVF Conference on Computer Vision and Pattern
  Recognition (CVPR)}, pages 4734--4743, 2022.

\bibitem{guo2020single}
Zichao Guo, Xiangyu Zhang, Haoyuan Mu, Wen Heng, Zechun Liu, Yichen Wei, and
  Jian Sun.
\newblock Single path one-shot neural architecture search with uniform
  sampling.
\newblock In {\em European Conference on Computer Vision}, 2019.

\bibitem{Habi2020HMQHF}
Hai~Victor Habi, Roy~H. Jennings, and Arnon Netzer.
\newblock Hmq: Hardware friendly mixed precision quantization block for cnns.
\newblock {\em ArXiv}, abs/2007.09952, 2020.

\bibitem{han2015deep}
Song Han, Huizi Mao, and William~J Dally.
\newblock Deep compression: Compressing deep neural networks with pruning,
  trained quantization and huffman coding.
\newblock {\em International Conference on Learning Representations}, 2016.

\bibitem{resnet}
Kaiming He, Xiangyu Zhang, Shaoqing Ren, and Jian Sun.
\newblock Deep residual learning for image recognition.
\newblock In {\em CVPR}, 2016.

\bibitem{li2021nas}
Yiming Hu, Xingang Wang, Lujun Li, and Qingyi Gu.
\newblock Improving one-shot nas with shrinking-and-expanding supernet.
\newblock {\em Pattern Recognition}, 2021.

\bibitem{jacob2018quantization}
Benoit Jacob, Skirmantas Kligys, Bo Chen, Menglong Zhu, Matthew Tang, Andrew
  Howard, Hartwig Adam, and Dmitry Kalenichenko.
\newblock Quantization and training of neural networks for efficient
  integer-arithmetic-only inference.
\newblock In {\em Proceedings of the IEEE conference on computer vision and
  pattern recognition}, pages 2704--2713, 2018.

\bibitem{Jin2019AdaBitsNN}
Qing Jin, Linjie Yang, and Zhenyu~A. Liao.
\newblock Adabits: Neural network quantization with adaptive bit-widths.
\newblock {\em 2020 IEEE/CVF Conference on Computer Vision and Pattern
  Recognition (CVPR)}, pages 2143--2153, 2019.

\bibitem{krishnamoorthi2018quantizing}
Raghuraman Krishnamoorthi.
\newblock Quantizing deep convolutional networks for efficient inference: A
  whitepaper.
\newblock {\em arXiv preprint arXiv:1806.08342}, 2018.

\bibitem{ref01_alexnet}
Alex Krizhevsky, Ilya Sutskever, and Geoffrey~E Hinton.
\newblock Imagenet classification with deep convolutional neural networks.
\newblock {\em NeurIPS}, 2012.

\bibitem{Lee2018SNIPSN}
Namhoon Lee, Thalaiyasingam Ajanthan, and Philip H.~S. Torr.
\newblock Snip: Single-shot network pruning based on connection sensitivity.
\newblock {\em ArXiv}, abs/1810.02340, 2018.

\bibitem{leman1968reduction}
AA Leman and Boris Weisfeiler.
\newblock A reduction of a graph to a canonical form and an algebra arising
  during this reduction.
\newblock {\em Nauchno-Technicheskaya Informatsiya}, 2(9):12--16, 1968.

\bibitem{Li2021AutoLossZeroSL}
Hao Li, Tianwen Fu, Jifeng Dai, Hongsheng Li, Gao Huang, and Xizhou Zhu.
\newblock Autoloss-zero: Searching loss functions from scratch for generic
  tasks.
\newblock {\em 2022 IEEE/CVF Conference on Computer Vision and Pattern
  Recognition (CVPR)}, pages 999--1008, 2021.

\bibitem{li2022self}
Lujun Li.
\newblock Self-regulated feature learning via teacher-free feature
  distillation.
\newblock In {\em ECCV}, 2022.

\bibitem{li2023auto}
Lujun Li, Peijie Dong, Zimian Wei, and Ya Yang.
\newblock Automated knowledge distillation via monte carlo tree search.
\newblock In {\em ICCV}, 2023.

\bibitem{li2022tf}
Lujun Li, Liang Shiuan-Ni, Ya Yang, and Zhe Jin.
\newblock Teacher-free distillation via regularizing intermediate
  representation.
\newblock In {\em IJCNN}, 2022.

\bibitem{Li2021BRECQPT}
Yuhang Li, Ruihao Gong, Xu Tan, Yang Yang, Peng Hu, Qi Zhang, Fengwei Yu, Wei
  Wang, and Shi Gu.
\newblock Brecq: Pushing the limit of post-training quantization by block
  reconstruction.
\newblock {\em ArXiv}, abs/2102.05426, 2021.

\bibitem{ZenNAS}
Ming Lin, Pichao Wang, Zhenhong Sun, Hesen Chen, Xiuyu Sun, Qi Qian, Hao Li,
  and Rong Jin.
\newblock Zen-nas: A zero-shot nas for high-performance image recognition.
\newblock {\em ICCV}, 2021.

\bibitem{lin2014mscoco}
Tsung{-}Yi Lin, Michael Maire, Serge~J. Belongie, James Hays, Pietro Perona,
  Deva Ramanan, Piotr Doll{\'{a}}r, and C.~Lawrence Zitnick.
\newblock Microsoft {COCO:} common objects in context.
\newblock In {\em Computer Vision - {ECCV} 2014 - 13th European Conference,
  Zurich, Switzerland, September 6-12, 2014, Proceedings, Part {V}}, pages
  740--755, 2014.

\bibitem{liu2018darts}
Hanxiao Liu, Karen Simonyan, and Yiming Yang.
\newblock Darts: Differentiable architecture search.
\newblock {\em ArXiv}, abs/1806.09055, 2018.

\bibitem{ickd}
Li Liu, Qinwen Huang, Sihao Lin, Hongwei Xie, Bing Wang, Xiaojun Chang, and
  Xiao-Xue Liang.
\newblock Exploring inter-channel correlation for diversity-preserved knowledge
  distillation.
\newblock {\em 2021 ICCV}, 2021.

\bibitem{Liu2021LossFD}
Peidong Liu, Gengwei Zhang, Bochao Wang, Hang Xu, Xiaodan Liang, Yong Jiang,
  and Zhenguo Li.
\newblock Loss function discovery for object detection via
  convergence-simulation driven search.
\newblock {\em ArXiv}, abs/2102.04700, 2021.

\bibitem{liu2021selfish}
Shiwei Liu, Decebal~Constantin Mocanu, Yulong Pei, and Mykola Pechenizkiy.
\newblock Selfish sparse rnn training.
\newblock {\em arXiv preprint arXiv:2101.09048}, 2021.

\bibitem{liu2023norm}
Xiaolong Liu, Lujun Li, Chao Li, and Anbang Yao.
\newblock Norm: Knowledge distillation via n-to-one representation matching.
\newblock In {\em ICLR}, 2023.

\bibitem{Lopes2021EPENASEP}
Vasco Lopes, Saeid Alirezazadeh, and Lu{\'i}s~A. Alexandre.
\newblock Epe-nas: Efficient performance estimation without training for neural
  architecture search.
\newblock In {\em ICANN}, 2021.

\bibitem{lou2019autoq}
Qian Lou, Feng Guo, Lantao Liu, Minje Kim, and Lei Jiang.
\newblock Autoq: Automated kernel-wise neural network quantization.
\newblock {\em arXiv preprint arXiv:1902.05690}, 2019.

\bibitem{Ma2021OMPQOM}
Yuexiao Ma, Taisong Jin, Xiawu Zheng, Yan Wang, Huixia Li, Guannan Jiang, Wei
  Zhang, and Rongrong Ji.
\newblock Ompq: Orthogonal mixed precision quantization.
\newblock {\em ArXiv}, abs/2109.07865, 2021.

\bibitem{mellor2020neural}
Joseph {Mellor}, Jack {Turner}, Amos~J. {Storkey}, and Elliot~J. {Crowley}.
\newblock Neural architecture search without training.
\newblock {\em arXiv preprint arXiv:2006.04647}, 2020.

\bibitem{morgan1991experimental}
Nelson Morgan et~al.
\newblock Experimental determination of precision requirements for
  back-propagation training of artificial neural networks.
\newblock In {\em Proc. Second Int’l. Conf. Microelectronics for Neural
  Networks}, pages 9--16. Citeseer, 1991.

\bibitem{Mozer1988SkeletonizationAT}
Michael~C. Mozer and Paul Smolensky.
\newblock Skeletonization: A technique for trimming the fat from a network via
  relevance assessment.
\newblock In {\em NIPS}, 1988.

\bibitem{Norwich1993InformationSA}
Kenneth~H Norwich.
\newblock {\em Information, sensation, and perception}.
\newblock Academic Press San Diego, 1993.

\bibitem{park2018value}
Eunhyeok Park, Sungjoo Yoo, and Peter Vajda.
\newblock Value-aware quantization for training and inference of neural
  networks.
\newblock In {\em Proceedings of the European Conference on Computer Vision
  (ECCV)}, pages 580--595, 2018.

\bibitem{Park2018ValueawareQF}
Eunhyeok Park, Sungjoo Yoo, and P{\'e}ter Vajda.
\newblock Value-aware quantization for training and inference of neural
  networks.
\newblock {\em ArXiv}, abs/1804.07802, 2018.

\bibitem{Real2020AutoMLZeroEM}
Esteban Real, Chen Liang, David~R. So, and Quoc~V. Le.
\newblock Automl-zero: Evolving machine learning algorithms from scratch.
\newblock In {\em International Conference on Machine Learning}, 2020.

\bibitem{shao2023catch}
Shitong Shao, Xu Dai, Shouyi Yin, Lujun Li, Huanran Chen, and Yang Hu.
\newblock Catch-up distillation: You only need to train once for accelerating
  sampling.
\newblock {\em arXiv preprint arXiv:2305.10769}, 2023.

\bibitem{qescore}
Zhenhong Sun, Ce Ge, Junyan Wang, Ming Lin, Hesen Chen, Hao Li, and Xiuyu Sun.
\newblock Entropy-driven mixed-precision quantization for deep network design
  on iot devices.
\newblock In {\em Advances in Neural Information Processing Systems}, 2022.

\bibitem{Sun2021MAEDETRM}
Zhenhong Sun, Ming Lin, Xiuyu Sun, Zhiyu Tan, Hao Li, and Rong Jin.
\newblock Mae-det: Revisiting maximum entropy principle in zero-shot nas for
  efficient object detection.
\newblock In {\em International Conference on Machine Learning}, 2021.

\bibitem{syflow}
Hidenori Tanaka, Daniel Kunin, Daniel~L Yamins, and Surya Ganguli.
\newblock Pruning neural networks without any data by iteratively conserving
  synaptic flow.
\newblock In {\em NeurIPS}, 2020.

\bibitem{Tanaka2020PruningNN}
Hidenori Tanaka, Daniel Kunin, Daniel L.~K. Yamins, and Surya Ganguli.
\newblock Pruning neural networks without any data by iteratively conserving
  synaptic flow.
\newblock {\em ArXiv}, abs/2006.05467, 2020.

\bibitem{Theis2018FasterGP}
Lucas Theis, Iryna Korshunova, Alykhan Tejani, and Ferenc Husz{\'a}r.
\newblock Faster gaze prediction with dense networks and fisher pruning.
\newblock {\em ArXiv}, abs/1801.05787, 2018.

\bibitem{Wang2020PickingWT}
Chaoqi Wang, ChaoQi Wang, Guodong Zhang, and Roger~Baker Grosse.
\newblock Picking winning tickets before training by preserving gradient flow.
\newblock {\em ArXiv}, abs/2002.07376, 2020.

\bibitem{wang2019haq}
Kuan Wang, Zhijian Liu, Yujun Lin, Ji Lin, and Song Han.
\newblock Haq: Hardware-aware automated quantization with mixed precision.
\newblock In {\em Proceedings of the IEEE/CVF Conference on Computer Vision and
  Pattern Recognition}, pages 8612--8620, 2019.

\bibitem{Wang2020APQ}
Tianzhe Wang, Kuan Wang, Han Cai, Ji Lin, Zhijian Liu, and Song Han.
\newblock Apq: Joint search for nerwork architecture, pruning and quantization
  policy.
\newblock In {\em Proceedings of the IEEE/CVF Conference on Computer Vision and
  Pattern Recognition}, 2020.

\bibitem{wu2018dnas}
Bichen Wu, Yanghan Wang, Peizhao Zhang, Yuandong Tian, Peter Vajda, and Kurt
  Keutzer.
\newblock Mixed precision quantization of convnets via differentiable neural
  architecture search.
\newblock {\em arXiv preprint arXiv:1812.00090}, 2018.

\bibitem{Yang2020FracBitsMP}
Linjie Yang and Qing Jin.
\newblock Fracbits: Mixed precision quantization via fractional bit-widths.
\newblock In {\em AAAI Conference on Artificial Intelligence}, 2020.

\bibitem{yao2021hawq}
Zhewei Yao, Zhen Dong, Zhangcheng Zheng, Amir Gholami, Jiali Yu, Eric Tan,
  Leyuan Wang, Qijing Huang, Yida Wang, Michael Mahoney, et~al.
\newblock Hawq-v3: Dyadic neural network quantization.
\newblock In {\em International Conference on Machine Learning}, pages
  11875--11886. PMLR, 2021.

\bibitem{Yao2020HAWQV3DN}
Zhewei Yao, Zhen Dong, Zhangcheng Zheng, Amir Gholami, Jiali Yu, Eric Tan,
  Leyuan Wang, Qijing Huang, Yida Wang, Michael~W. Mahoney, and Kurt Keutzer.
\newblock Hawqv3: Dyadic neural network quantization.
\newblock In {\em International Conference on Machine Learning}, 2020.

\bibitem{yu2020search}
Haibao Yu, Qi Han, Jianbo Li, Jianping Shi, Guangliang Cheng, and Bin Fan.
\newblock Search what you want: Barrier panelty nas for mixed precision
  quantization.
\newblock In {\em European Conference on Computer Vision}, pages 1--16.
  Springer, 2020.

\bibitem{zhang2018lq}
Dongqing Zhang, Jiaolong Yang, Dongqiangzi Ye, and Gang Hua.
\newblock Lq-nets: Learned quantization for highly accurate and compact deep
  neural networks.
\newblock In {\em Proceedings of the European conference on computer vision
  (ECCV)}, pages 365--382, 2018.

\end{thebibliography}
}

\end{document}